\RequirePackage{fix-cm}
\documentclass[twocolumn,natbib]{svjour3_nomanuscriptname}          % twocolumn
\smartqed  % flush right qed marks, e.g. at end of proof
\usepackage{graphicx}

\usepackage{epsfig}
\usepackage{amsmath}
\usepackage{amssymb}

\usepackage{pifont}% http://ctan.org/pkg/pifont
\newcommand{\cmark}{\ding{51}}%
\newcommand{\xmark}{\ding{55}}%
\usepackage{url}
\usepackage{multirow}
\usepackage[pagebackref=true,breaklinks=true,letterpaper=true,colorlinks,bookmarks=false]{hyperref}

\usepackage{xspace}
\makeatletter
\DeclareRobustCommand\onedot{\futurelet\@let@token\@onedot}
\def\@onedot{\ifx\@let@token.\else.\null\fi\xspace}
\def\eg{\emph{e.g}\onedot} 
\def\ie{\emph{i.e}\onedot} 
 
\def\etc{\emph{etc}\onedot}
\def\vs{\emph{vs}\onedot}

\makeatother

\begin{document}

\hyphenation{Mimetics}

%\title{Insert your title here%\thanks{Grants or other notes
\title{Mimetics: Towards Understanding Human Actions Out of Context}



%\author{First Author         \and
%        Second Author %etc.
%}

\author{Philippe Weinzaepfel \and Gr\'egory Rogez}


% \institute{F. Author \at
%               first address \\
%               Tel.: +123-45-678910\\
%               Fax: +123-45-678910\\
%               \email{fauthor@example.com}           %  \\
% %             \emph{Present address:} of F. Author  %  if needed
%            \and
%            S. Author \at
%               second address
% }

\institute{NAVER Labs Europe \\
	      6 chemin de Maupertuis, 38240 Meylan, France \\
              Tel.: +33-476-615-050\\
              \email{firstname.lastname@naverlabs.com} %\\
}

\date{}

\maketitle

%%%%%%%%% ABSTRACT
\begin{abstract}
Recent methods for video action recognition have reached outstanding performances on existing benchmarks.
However, they tend to leverage context such as scenes or objects instead of focusing on understanding the human action itself.
For instance, a tennis field leads to the prediction \emph{playing tennis} irrespectively of the actions performed in the video.
In contrast, humans have a more complete understanding of actions and can recognize them without context.
The best example of out-of-context actions are mimes, that people can typically recognize despite missing relevant objects and scenes.
In this paper, we propose to benchmark action recognition methods in such absence of context 
and  introduce a novel dataset, \emph{Mimetics}, consisting of mimed actions for a subset of 50 classes from the Kinetics benchmark.
Our experiments show that (a) state-of-the-art 3D convolutional neural networks obtain disappointing results on such videos, highlighting the lack of true understanding of the human actions and (b) models leveraging body language via human pose are less prone to context biases. In particular, we show that applying a shallow neural network with a single temporal convolution over body pose features transferred to the action recognition problem performs surprisingly well compared to 3D action recognition methods.

\end{abstract}

%%%%%%%%% BODY TEXT
\section{Introduction}

\begin{figure}
 \centering
 \includegraphics[width=\linewidth]{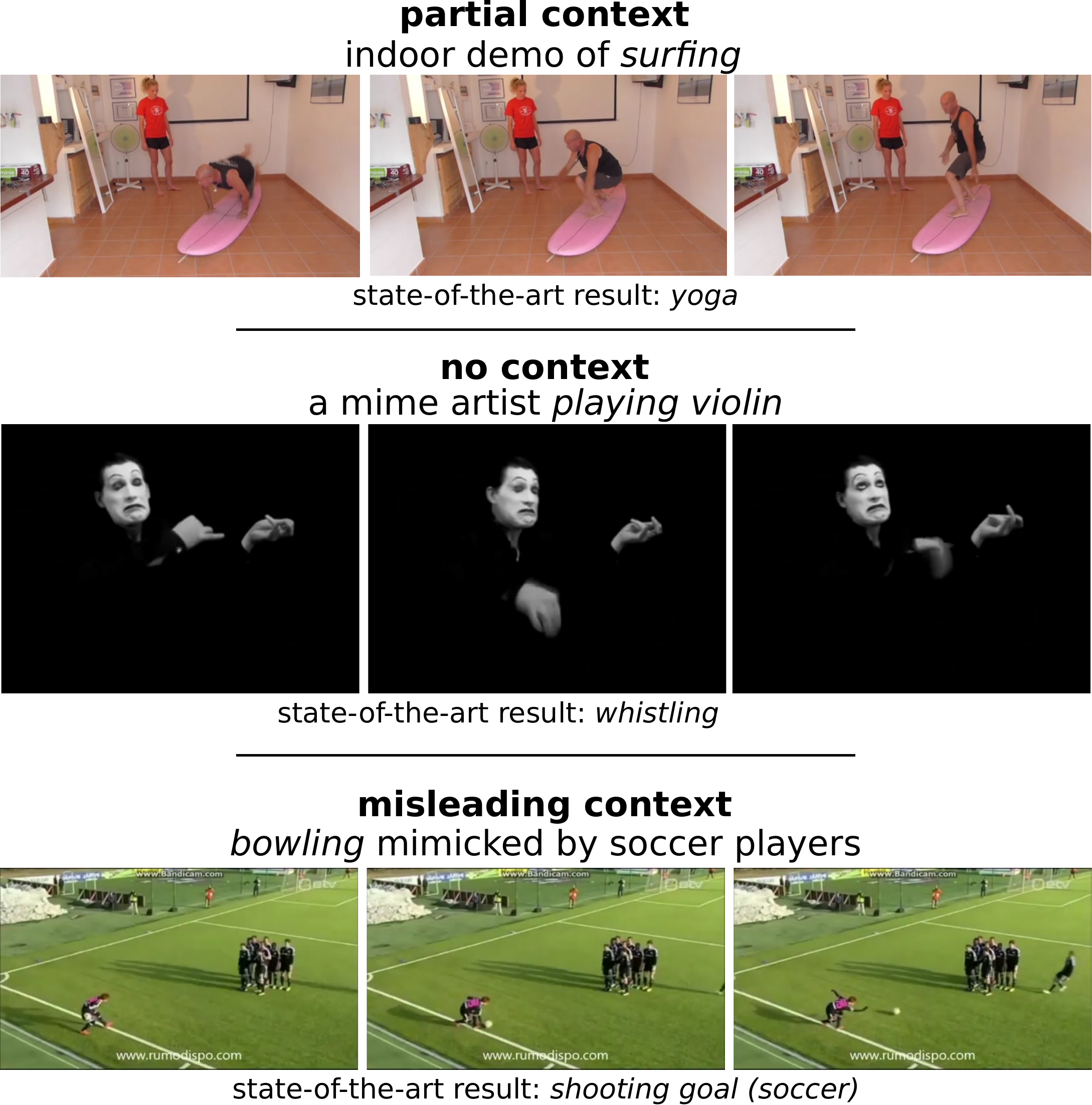}
 \caption{Examples where context is partial (top), absent (middle), or  misleading (bottom). The first row shows someone training indoor for \textit{surfing}, with the right object but not in a standard place. The second row shows a mime artist mimicking someone \textit{playing violin}, but the object and the scene are absent. The third example contain a misleading context: soccer players are mimicking a scene of \textit{bowling} on a soccer field with a soccer ball. In all these cases, state-of-the-art 3D CNNs fail to recognize the actions.}
 \label{fig:splash}
\end{figure}
Action recognition has made remarkable progress over the past few years~\citep{I3D,slowfast,TwoStream,TSN}.
Most state-of-the-art methods~\citep{I3D,hara2018can,R2+1D} are built upon deep spatio-temporal convolutional achitectures applied on short clips of RGB frames.
These approaches achieved impressive classification performance, with a top-1 accuracy over 77\% on the Kinetics dataset \citep{Kinetics}, and a top-5 accuracy of more than 93\%.
However, the explanations behind such performances remain unclear.
In particular, recent works~\citep{resound,repair, JacquotYK20} have shown that most datasets, and thus what Convolutional Neural Networks (CNNs) learn, are biased by static context such as scenes and objects.
For instance, Figure~\ref{fig:splash} shows some examples where context is only partial, absent or misleading, and that are misclassified by state-of-the-art 3D CNNs.
In particular, the last video taking place on a soccer field is classified as \textit{shooting goal (soccer)}, regardless of the actual action performed in the video.

\begin{figure}
 \centering
 \includegraphics[width=0.49\linewidth]{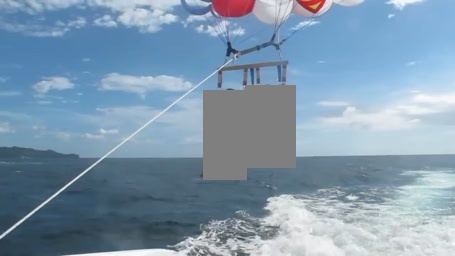}
 \hfill 
 \includegraphics[width=0.49\linewidth]{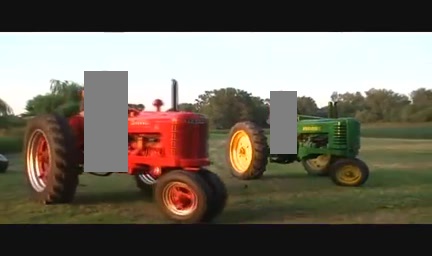}
 \caption{Examples of training frames where humans are masked. The context clearly suffices to guess the action \textit{parasailing} (left) and \textit{driving tractor} (right).}
 \label{fig:grey}
\end{figure}

To further assess the bias of existing datasets towards scenes and objects, we retrain a model on Kinetics after masking out all the humans in the videos, see Figure~\ref{fig:grey}.
The performance of this model on the original test set is around 65\%, which is extremely high for a model that has never seen any human at training.
This shows that scenes and objects are often sufficient to correctly classify the actions.

While this contextual information is certainly useful to predict human actions, it is not sufficient to truly understand what is happening in a scene. 
Humans have a more complete understanding of actions and can even recognize them without any context, object or scene. 
The most obvious example is given by mime artists, see middle row of Figure~\ref{fig:splash}, who can suggest emotions or actions to the audience using only facial expressions, gestures and movements, but without words or context. 
Mime as an art originates from ancient Greece and reached its heights with sixteen century Commedia dell'Arte,
 but it is considered one of the earliest mediums of expressions even before the appearance of spoken language. 
We claim that an intelligent system should also be able to understand mimed actions.

To understand action in out-of-context scenarios, \ie, when object and scene are absent or
misleading as shown in Figure~\ref{fig:splash}, action recognition can only rely on \emph{body language} captured by human pose and motion.
Such a cue is leveraged in the well-established field of 3D skeleton-based action recognition, also called \emph{3D action recognition}~\citep{du2015hierarchical,liu2016spatio,zhu2016co}, 
that take as input sequences of 3D pose skeletons. These methods have shown impressive results, 
validating that contextual information is not always necessary to recognize actions.
However, they are usually trained and tested on accurate and scripted sequences of 3D human poses, 
captured with RGB-D sensor~\citep{NTU} or Motion Capture systems~\citep{du2015hierarchical,zhu2016co} in constrained and unrealistic environments. 
To the best of our knowledge, 3D action recognition has never been applied to real-world situations and videos captured in the wild.
Another of our contributions is therefore to study whether such techniques generalize in-the-wild
 given current pose detectors and can be employed in out-of-context scenarios.

Recent human pose estimation methods~\citep{VNect,LCRNet++}, allow to estimate 3D poses of multiple people from a single image.
In this paper, we employ LCR-Net++~\citep{LCRNet++} to extract human 3D pose information from videos. It has shown robustness to challenging cases like occlusions and truncations by image boundary, estimating full-body 2D and 3D poses for every person in an image.
We compare three different action recognition baselines based on these poses.
The most intuitive pipeline is to detect 3D human poses in every frame, 
build 3D pose sequences by linking detections over time, and apply a state-of-the-art 3D action recognition algorithm.
However, such a method is likely to be sensitive to the level of noise inherent to 3D pose estimation in the wild.
The second baseline applies graph convolutions on 2D pose sequences, without 3D information, which might have the advantage to be more accurate.
We finally study another approach where 1D temporal convolutions are applied on human-level intermediate pose feature representations from LCR-Net++.
In other words, we transfer the features learned for 2D-3D pose estimation to action recognition: they typically contain information about the human poses without explicitly representing them as body keypoint coordinates.

Finally, to benchmark action recognition methods in out-of-context scenarios, we introduce the \textbf{Mimetics} dataset\footnote{\url{https://europe.naverlabs.com/research/computer-vision/mimetics/}}.
It contains over 700 video clips of mimed actions for a subset of 50 classes from the Kinetics dataset.
Mimetics allows to evaluate on mimed actions models that have been trained on Kinetics.
It is not meant to be used as training data. Our claim is that systems that supposedly try to reach human performance should be able to recognize actions out of context as
humans do without seeing mimes at training.
For further analysis, we additionally annotate for each clip whether an object gives clues on the action or not, and similarly for the scene. 
We also labelled the size of the objects for a fine-grained analysis of the bias towards objects.
We evaluate a state-of-the-art 3D convolutional network, and confirm that these models are biased towards scenes and objects.
Pose-based action recognition provides a more interpretable output but can lack fine-grained pose details, \eg, face and hands, for higher performance.

This paper is organized as follows. 
After reviewing related work in Section~\ref{sec:related}, we study the  bias of state-of-the-art action recognition datasets and models in Section~\ref{sec:bias}. 
Section~\ref{sec:method} then presents various pose-based baselines and compares them on existing action recognition datasets. 
Finally, Section~\ref{sec:mimes} introduces the Mimetics dataset and analyzes the performance on out-of-context action recognition.

\section{Related work}
\label{sec:related}

We benchmark action recognition approaches, comparing standard CNNs on RGB clips with pose-based methods.
This latter category can be further split into 2D pose-based approaches and 3D action recognition.

\noindent \textbf{Action classification in real-world videos.}
Different strategies have been deployed to handle video processing with CNNs such as two-stream architectures~\citep{FeichtenhoferConvolutional,TwoStream}, 
 Recurrent Neural Networks (RNNs)~\citep{Donahue_LSTM}, or spatio-temporal 3D convolutions~\citep{I3D,slowfast,C3D}.
\cite{TwoStream} introduced a two-stream architecture with 2D convolutions, 
in which one stream captures appearance information from RGB inputs while the second one operates on optical flow representation and models motion.
While improvements of this approach have been proposed~\citep{FeichtenhoferConvolutional}, 
most state-of-the-art methods now use a 3D deep convolutional network~\citep{I3D,C3D,R2+1D,XieRethinking}, optionally in combination with a two-stream architecture.
Compared to 2D convolutions, 3D convolutions allow to leverage spatio-temporal information at the cost of a higher number of parameters and higher computational cost.
With recent very large-scale datasets such as Kinetics~\citep{Kinetics}, it is possible to train such 3D CNNs effectively~\citep{hara2018can}, 
and impressive performances can be obtained even on small datasets, thanks to pretraining on Kinetics~\citep{I3D}.
For instance, I3D~\citep{I3D} achieved state-of-the-art accuracy on HMDB51~\citep{HMDB} and UCF101~\citep{UCF101} using a two-stream network with a 3D Inception backbone~\cite{Inception}.
\cite{R2+1D} and \cite{XieRethinking} replaced 3D convolutions with separate spatial and temporal convolutions, which reduces the number of parameters to learn.
However, all these methods lack a clear understanding of their classification choices. 
In particular, recent studies~\citep{resound,repair} suggest that they tend to leverage dataset biases instead of focusing on the human action.

\noindent \textbf{2D pose for action classification in real-world videos.}
An insightful diagnostic to understand what affects the action recognition results most was provided by \cite{JHMDB}, who found that high-level 2D pose features greatly outperform low/mid level features.
This has motivated further research on incorporating 2D body poses information in real-world action recognition models~\citep{ActionXPose,PCNN,rpan,iqbal2017pose,StarNet,ActionMachine}.
For instance, this can be done by pooling features~\citep{cao2016action,PCNN} or defining an attention mechanism~\citep{rpan,girdharNIPS17}. 
However, this leads to limited gain and often assumes that humans are fully-visible.
\cite{ChainedMS} trained a 3D CNN on human part segmentation inputs, and added a third stream to two-stream networks.
Some other recent methods have shown improved action recognition performance by incorporating 2D pose information from off-the-shelf pose detectors~\citep{PoTion,liu2018recognizing,wang2018pose}.
For instance, \cite{PoTion} and \cite{liu2018recognizing} extract joint heatmaps and encode their evolution over time. 
\cite{wang2018pose} define a two-stream network: one stream encodes the evolution of the pose while the second one models relationship with objects. 
However, it remains limited to single-person action recognition.
\cite{luvizon20182d} propose a multi-task architecture where 2D poses are predicted at the same time as appearance features are pooled over body joints for action recognition.

\noindent \textbf{3D action recognition.}
Compared to 2D poses, 3D poses have the advantage to be unambiguous and to better handle motion dynamics. 
Recent attempts on 3D action recognition have employed RNNs to handle sequential data and to model the contextual dependencies in the temporal domain~\citep{du2015hierarchical,liu2016spatio,si2018skeleton,weng2018deformable,zhu2016co}. 
\cite{du2015hierarchical} propose a hierarchical RNN in which the human skeleton was divided into five parts (arms, legs and trunk) 
to feed five different subnets later fused hierarchically. 
\cite{zhu2016co} added a mixed-norm regularization term to a RNN cost function in order to learn the co-occurrence features of skeleton joints for action classification. 
More recently, simple CNN-based methods applied to the 2D or 3D joint coordinates have shown to outperform more complex RNN architectures~\citep{du2015skeleton}.
In a similar spirit, \cite{STGCN} represent the sequence of poses as a graph, and apply a spatio-temporal graph convolutional network (STGCN) to recognize actions.
Most of these algorithms use 3D human poses obtained from a Motion Capture system~\citep{du2015hierarchical,zhu2016co}, a Kinect sensor~\citep{liu2016spatio} or a multi-camera setting~\citep{yao2012coupled}, and none of them experimented on real-world videos with estimated 3D poses.

To the best of our knowledge, we are the first to analyze 3D action recognition in real-world videos.
\cite{STGCN} show that their STGCN method can also be applied in the wild, but they only use 2D poses in this scenario.
More precisely, they extract 2D human poses with OpenPose~\citep{OpenPose}, build a graph using the 2 highest-scored detections per frame, and apply their spatio-temporal graph network, replacing X,Y,Z coordinates of the 3D poses, by $x,y,s$, where $x,y$ are the 2D coordinates of the joint, normalized into $[-0.5,0.5]$ and $s$ is the score for this keypoint.
In the framework of \cite{luvizon20182d}, the multi-task architecture can deal with 2D and 3D poses at the same time as action recognition. 
However, ground-truth keypoints are required for training, and the 3D component is disabled for datasets in-the-wild, \ie, without 3D ground-truth poses.

\section{Context biases in action recognition}
\label{sec:bias}

To assess how much context is leveraged by current methods based on spatio-temporal CNNs, we consider videos where people are masked out.
To do so, we extracted human tubes in all videos using LCR-Net++~\citep{LCRNet++} detections linked over time (see Section~\ref{sub:tubes} for a detailed description) and removed all the humans from the video frames by coloring the tubes content in grey, see Figure~\ref{fig:grey}.

We performed this experiment on the standard Kinetics dataset~\citep{Kinetics} which consists of around 240k training videos, 20k for validation and 40k for testing for a total of 400 classes.
As a state-of-the-art model, we use a 3D CNN model, \ie, with spatio-temporal convolutions instead of 2D convolutions, using a ResNeXt-101 backbone~\citep{resnext}.
We first evaluate a 3D CNN trained on original videos and tested on masked videos, thus measuring the biases learned by the model.
Mean top-1 accuracy on the validation set is reported in Table~\ref{tab:grey}. 
It remains close to 40\%, which is extremely high given that there is no human from which the action can be recognized in the test videos.
This prediction is thus based on the remaining content of the video, \ie, context such as objects or scenes.

\begin{table}
 \caption{Mean top-1 accuracy (in \%) when training and testing a 3D CNN model on Kinetics using the original videos, videos where humans are masked (tubes maked) or videos where areas outside human tubes are masked (background masked)}
 \centering
 \resizebox{\linewidth}{!}{
 \begin{tabular}{cc|ccc}
  &  & \multicolumn{3}{c}{test on} \\
  &  & original & masked tubes  & masked background \\
  \hline
\multirow{3}{*}{train on}&   original & 74.5 & 38.7 & 45.7 \\
  & masked tubes & 65.7 & 63.9 & 41.8 \\
  & masked background & 65.8 & 45.7 & 65.3 \\
  \hline
 \end{tabular}
 }
 \label{tab:grey}
\end{table}

To better measure the biases of the dataset itself, we have trained a 3D CNN model on the masked videos and obtain 65.7\% on the original videos, down by only 8.8\% compared to training on the original data.
This performance is outstanding for a model that has not seen any human during training, and therefore has not really seen any action.
To further analyze this aspect, we additionally show in Table~\ref{tab:greycls} the classes with the most increase of accuracy.
Masking the actors at training increases the accuracy for classes in which the scene context (\eg \emph{long jump}, \emph{playing basketball}) or the presence of large objects (\eg \emph{driving tractor}) are sufficient to recognize the actions, see also Figure~\ref{fig:grey}.

\begin{table}
\caption{Classes with the most increase in accuracy (in \%) on Kinetics validation set when training on original videos or masked videos. The last column highlights the difference between these two settings.}
 \centering
 \footnotesize{}{}{
 \begin{tabular}{c|rr|r}
 class & original & masked & diff. \\
 \hline
building shed                  &   74.5 &   85.1 &  +10.6 \\
long jump                      &   62.0 &   72.0 &  +10.0 \\
driving tractor                &   68.1 &   76.6 &   +8.5 \\
riding elephant                &   86.0 &   94.0 &   +8.0 \\
tying knot (not on a tie)      &   64.0 &   72.0 &   +8.0 \\
playing basketball             &   66.0 &   74.0 &   +8.0 \\
changing oil                   &   85.7 &   91.8 &   +6.1 \\
planting trees                 &   77.6 &   83.7 &   +6.1 \\
peeling potatoes               &   59.2 &   65.3 &   +6.1 \\
parasailing                    &   82.0 &   88.0 &   +6.0 \\
\hline
 \end{tabular}
}
\label{tab:greycls} 
\end{table}

Such bias problem can be tackled by sampling over multiple datasets or reweighting samples, as shown for action~\citep{resound,repair} or object recognition~\citep{hyojin2019arxiv,undoing,unbiased}. 
For action recognition, another direction is to leverage body language which is not affected by this context bias.

We also consider the problem where all areas outside the humans are used for training and/or testing. 
When evaluating on these data the model trained on the original images, we still obtain a performance of 45.7\%.
When training a model on such videos and evaluating it on the same kind of data or on the original videos, a performance of around 65\% is obtained.  
Note that, despite masking the background areas, there are still lot of context information in the bounding boxes containing the humans: for instance they are still lot of snow in the box area when filming someone \textit{skiing slalom}, or objects that are used by a person often appear near that person and masking the areas outside the human tubes will not necessary lead to a model focusing uniquely on humans.

\section{Real-world 3D action recognition baselines}
\label{sec:method}

We benchmark three baselines, that all require the extraction of human tubes (Section~\ref{sub:tubes}).
We present two different methods that employ a spatio-temporal graph convolutional network, on explicit 3D (Section~\ref{sub:3D}) or
2D (Section~\ref{sub:2D}) pose sequences respectively.
Next, we introduce a third approach that consists of a single 1D temporal convolution applied on mid-level implicit pose features (Section~\ref{sub:feats}).
Finally, we present experimental results on existing benchmarks in Section~\ref{sub:xp}.

\subsection{Extracting human tubes}
\label{sub:tubes}

\noindent \textbf{Overview of LCR-Net++.}
We build our tube extraction and pose estimation upon LCR-Net++~\citep{LCRNet++}, which leverages a Faster R-CNN like architecture~\citep{Faster} with a ResNet-50 backbone~\citep{ResNet}.
A Region Proposal Network extracts candidate boxes around humans.
These regions are then classified into different so-called `anchor poses' that replace standard object classes: these key poses typically correspond to a person standing, a person sitting, \etc
Poses are then refined using a regression branch, that takes as input the same features used for classification.
Anchor-poses are defined jointly in 2D and 3D, and the refinement occurs in this joint 2D-3D pose space.
The detection framework allows to handle multiple people in a scene.
As the approach is holistic, it outputs full-body poses, even in case of occlusions or truncation by image boundaries.
We use the real-time model released by the authors\footnote{\url{http://thoth.inrialpes.fr/src/LCR-Net/}}, allowing experiments on large-scale datasets.

\noindent \textbf{Tube extraction.}
In order to leverage the evolution of poses over time, one needs to track each individual, \ie, to obtain human tubes~\citep{HumanTubes}.
We proceed by first running LCR-Net++ in every frame and follow standard procedures used in the spatio-temporal action localization literature~\citep{ACT,singh2017online} to link detections over time.
Starting from the highest scored detection, we match it with the detections in the next frame based on the Intersection-over-Union (IoU) between boxes. 
We link it if the IoU is over $0.3$. 
Otherwise, we match it to the frame after, and perform linear interpolation in the missing frames. 
We stop a tube if there was no match during $10$ consecutive frames. 
This procedure is run forward and backward to obtain a human tube.
We then delete all detections in this first link, and repeat the procedure for the remaining detections.
At training, we label the tubes with the video class. 
At test time, for each video and for each class, we take the maximum score over all tubes.

\begin{figure}
 \centering
 \includegraphics[width=\linewidth]{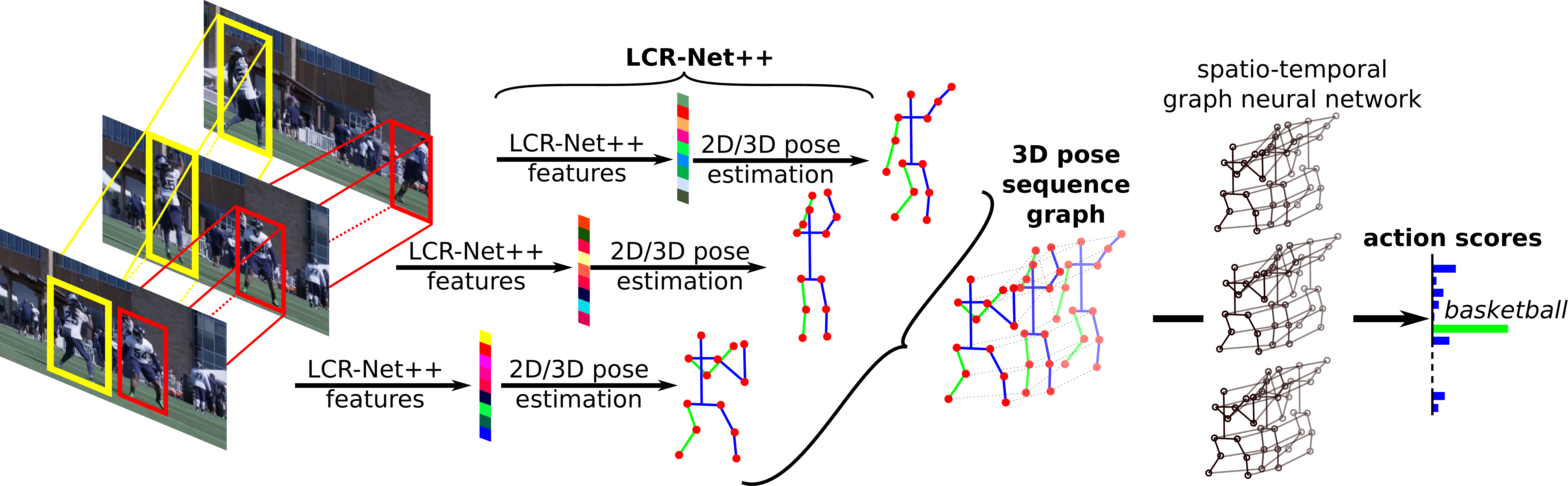}
 \caption{Overview of the STGCN3D baseline. Given an input video, we run LCR-Net++ to detect human tubes  (yellow and red boxes) and estimate 2D/3D poses (shown only for the yellow tube for readability).  We  build 3D pose sequences and run a state-of-the-art 3D action recognition method based on spatio-temporal graph neural network~\citep{STGCN} to obtain action scores.}
 \label{fig:3D}
\end{figure}

\subsection{Baseline based on explicit 3D pose}
\label{sub:3D}

Figure~\ref{fig:3D} shows an overview of the most intuitive baseline.
It is based on explicit 3D pose information.
More precisely, given the human tubes, we extract the 3D poses estimated by LCR-Net++ for each box, thus building a 3D pose skeleton sequence for each tube.
We finally run a state-of-the-art 3D action recognition method, namely STGCN, using the code released by \cite{STGCN}\footnote{\url{https://github.com/yysijie/st-gcn}}.
The idea consists in building a graph in space and time from the pose sequence, on which spatio-temporal convolution are applied.
We denote this first baseline as \textbf{STGCN3D}.

\subsection{Variant based on explicit 2D pose}
\label{sub:2D}

As the STGCN method of \cite{STGCN} has also been applied to 2D poses,
we use a variant of the previous pipeline, replacing the 3D poses estimated by LCR-Net++ by its 2D poses.
Note that  2D poses extracted by LCR-Net++ can partially fall outside the image boundaries as the method is holistic and always produces a full-body estimate, even when the person is occluded or truncated at image boundaries. In STGCN, joints coordinates are transformed from pixel coordinates to [-0.5,0.5], some joints estimated by LCR-net++ can therefore have absolute values slightly above 0.5.
On the one hand, this variant is likely to get worse performance, as 3D poses are more informative than 2D poses which are inherently ambiguous.
But on the other hand, 2D poses extracted from images and videos tend to be more accurate than 3D poses which are more prone to noise.
We call this second baseline \textbf{STGCN2D}.

\subsection{Temporal convolution on implicit pose features}
\label{sub:feats}

\begin{figure}
 \centering
 \includegraphics[width=0.85\linewidth]{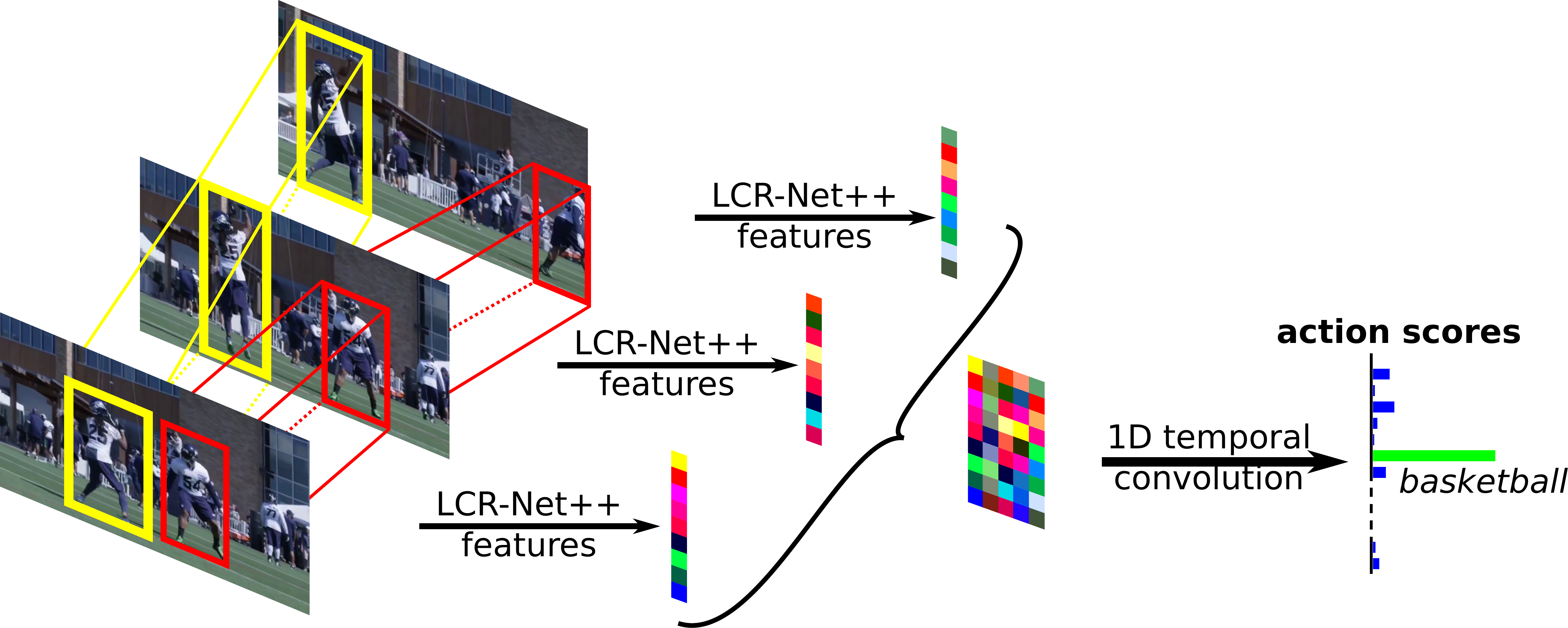}
 \caption{Overview of the implicit pose baseline, SIP-Net. Given an input video, we run LCR-Net++ to detect humans tubes (yellow and red boxes) and extract mid-level pose features (shown only for the yellow tube for readability). We stack them over time and apply a single 1D temporal convolution to obtain action scores.}
 \label{fig:method}
\end{figure}

We finally study a baseline that transfers the implicit pose representation carried by mid-level features within LCR-Net++, 
without using explicit body keypoint coordinates, see Figure~\ref{fig:method}.
We select the features used as input to the final layers for pose classification and refinement.
These features have 2048 dimensions with a ResNet50 backbone and carry information about both 2D and 3D poses.
The features  are stacked over time along human tubes and a temporal convolution of kernel size $T$ is applied on top of the resulting matrix.
This convolution outputs action scores for the sequence.

At training, we sample random clips of $T$ consecutive frames and use a cross-entropy loss.
At test time, we use a fully-convolutional architecture and average the class probabilities by a softmax on the scores for all clips in the videos.
We did experiment with deeper network on top of the stacked features but did not see any significant improvement.
Due to GPU memory constraint, we freeze the weights of LCR-Net++ during training, allowing larger temporal windows to be considered.
We denote this third baseline as \textbf{SIP-Net} for Stacked Implicit Pose Network.

\subsection{Comparison on existing datasets}
\label{sub:xp}

Before comparing these baselines on out-of-context actions (Section~\ref{sec:mimes}), we assess their performance for real-world action recognition on existing datasets, with various levels of ground-truth.
Table~\ref{tab:datasets} summarizes them in terms of number of videos, classes, splits, as well as frame-level ground-truths.
For datasets with multiple splits, some results are reported on the first split only, denoted for instance as JHMDB-1 for the split 1 of JHMDB.
While our goal is to perform action recognition in real-world videos, 
we validate the baselines on the constrained NTU 3D action recognition dataset~\citep{NTU} that contains ground-truth poses in 2D and 3D, using the standard cross-subject (cs) split.
We also experiment on the JHMDB~\citep{JHMDB} and PennAction~\citep{PennAction} datasets that have ground-truth 2D poses, but no 3D poses.
Finally, we use HMDB51~\citep{HMDB}, UCF101~\citep{UCF101} and Kinetics~\citep{Kinetics} that contain no more information than the ground-truth label of each video.
As metric, we report the standard mean accuracy, \ie, the ratio of correctly classified videos per class, averaged over all classes.

\begin{table}
\caption{Overview of the datasets used in our experiments}
\centering
\resizebox{\linewidth}{!}{
\begin{tabular}{l|rrc|ccc}
             & \small{}{}{\#cls} & \multicolumn{1}{c}{\small{}{}{\#vid}} & \small{}{}{\#splits} & \small{}{}{in-the-wild} & \small{}{}{GT 2D} & \small{}{}{GT 3D}  \\
\hline
NTU          &    60     &  56,578    &  2        & \xmark  & \cmark & \cmark\\
\hline
JHMDB        &    21     &     928    &  3        & \cmark & \cmark & \\
PennAction   &    15     &   2,326    &  1        & \cmark & \cmark & \\
HMDB51       &    51     &   6,766    &  3        & \cmark & & \\
UCF101       &   101     &  13,320    &  3        & \cmark & & \\
Kinetics     &   400     & 306,245    &  1        & \cmark & & \\
\hline
\end{tabular}
}
\label{tab:datasets}
\end{table}

In Appendix~\ref{app:xp}, we report various experiments based on this various levels of ground-truth, allowing to study the impact of extracted tubes, extracted poses as well as the benefit of transferring pose features for SIP-Net. We also plot the performance of SIP-Net with varying $T$ and use $T=32$ in the remaining of this work.

\begin{table*}
\caption{Mean accuracies (in \%) for our three baselines on all datasets, and for state-of-the-art pose-based approaches}
\centering
\resizebox{\linewidth}{!}{
\begin{tabular}{l|ccccccccc}
 & JHMDB-1 & JHMDB & PennAction & NTU (cs) & HMB51-1 & HMDB51 & UCF101-1 & UCF101 & Kinetics \\
 \hline
 \hline
 PoTion~\citep{PoTion}              &   59.1  &   57.0  &    -    &   -    &    46.3   &   43.7   &    60.5  &    65.2  &   16.6 \\
 \cite{ChainedMS} (pose only) & 45.5   &   -       &     -      &   67.8   &   36.0   &    -       &   56.9   &   -    &     -    \\
 MultiTask ~\citep{luvizon20182d}  (uses RGB) & - &     -    &    {\bf97.4  } &   74.3  &      -     &     -     &     -      &      -     &     -    \\
 STGCN~\citep{STGCN}  (OpenPose) & 25.2  & 25.4  & 71.6  &   {\bf79.8}   & 38.6   &  34.7   & 54.0  &   50.6   &  30.7  \\
 \hline
 STGCN2D                           &   23.2 &   23.2 &   85.5  &   69.4  &   36.5  &   32.7  &   49.2  &   44.4  &   11.9 \\
 STGCN3D                           &   53.1 &   50.5 &   89.2  &   75.0  &   39.8  &   41.0  &   48.5  &   51.1  &   10.6 \\
 SIP-Net                           &   {\bf66.4} &   {\bf62.4} &   93.5    &   64.8  & {\bf50.7}  & {\bf51.2}  &  {\bf66.1}  & {\bf66.0}  & {\bf32.8} \\
 \hline
\end{tabular}
}
\label{tab:LCR}
\end{table*}

Table~\ref{tab:LCR} provides a comparison of the mean accuracy on all datasets (last three rows).
The method based on implicit pose features (SIP-Net) significantly outperforms the baselines that employ explicit 2D and 3D poses, except on NTU.
The gap is over 10\% on HMDB51, UCF101 and Kinetics.
This can be explained by the fact that explicitly extracting the poses lead to a significant level of noise in the body keypoint representations for in-the-wild videos.
Using an implicit pose representation as in SIP-Net allows for more robustness.
Interestingly, on HMDB51, UCF101 and Kinetics, the 2D pose baseline performs slightly better than the 3D, suggesting that 3D pose suffers from much more noise in unconstrained videos.

Finally, we compare our baselines to the state of the art among pose-based methods, see Table~\ref{tab:LCR}.
SIP-Net obtains a higher accuracy than PoTion~\citep{PoTion} with a margin over 5\% on JHMDB, HMDB51 and UCF101-1, and  of 16\% on Kinetics.
Compared to the pose model only of \cite{ChainedMS}, we obtain a higher accuracy on JHMDB, HMDB51 and UCF101.
On NTU and PennAction, \cite{luvizon20182d} obtain a higher accuracy because their approach also leverages appearance features.
When combining SIP-Net with a standard RGB stream using 3D ResNeXt-101 backbone, we obtain 98.9\% on PennAction.
Finally, as in~\citep{STGCN}, we  run STGCN code on 2D poses detected by OpenPose~\citep{OpenPose}. 
We significantly outperform this approach on JHMDB, PennAction, HMDB51 and UCF101.
On Kinetics, the gap is much smaller, with only 2\%.
This dataset contains many videos with very near close-ups on faces or captured from a first-person viewpoint, which leads to a large number of misdetections by LCR-Net++ that has not been trained in such conditions.
For videos where only the face is visible, OpenPose that outputs 18 keypoints including 5 on the head (nose, two ears, two eyes) is able to detect a pose. 
In contrast, LCR-Net++ that estimates only 1 (out of 13) keypoint on the center of the head, fails to detect humans in such cases. 
Table~\ref{tab:ourskinetics} shows the 10 classes with the highest and lowest accuracy for SIP-Net.
Classes with high top-1 accuracy can be clearly recognized from body pose only.
In contrast, the classes at 0\% are either actions often captured in first-person viewpoint where the poses are not detected (\emph{making a cake}), or classes with no motion of the body keypoint as they mainly contain motion of the face (\emph{sniffing}) or the hands (\emph{drumming fingers}).

\begin{table}
\caption{Classes with the highest/lowest accuracy (in \%) for SIP-Net on Kinetics}
\centering
\resizebox{\linewidth}{!}{
\begin{tabular}{lr|lr}
\multicolumn{2}{c|}{highest top-1 accuracy} & \multicolumn{2}{c}{lowest top-1 accuracy} \\
\hline
crawling baby                       &   91.8  & rock scissors paper                 &    0.0 \\
presenting weather forecast         &   90.0  & throwing ball                       &    0.0 \\
riding mechanical bull              &   89.8  & eating chips                        &    0.0 \\
deadlifting                         &   88.9  & drumming fingers                    &    0.0 \\
surfing crowd                       &   87.5  & tossing coin                        &    0.0 \\
arm wrestling                       &   87.5  & sniffing                            &    0.0 \\
filling eyebrows                    &   84.4  & unloading truck                     &    0.0 \\
shearing sheep                      &   83.7  & holding snake                       &    0.0 \\
bench pressing                      &   82.0  & making a cake                       &    0.0 \\
front raises                        &   81.6  & ripping paper                       &    0.0 \\
\hline
\end{tabular}
}
\label{tab:ourskinetics} 
\end{table}

\section{Experiments on mimed actions}
\label{sec:mimes}

To assess the bias of action recognition algorithms towards scenes and objects, and evaluate their generalizability  in absence of such visual context, we introduce \textbf{Mimetics}, a dataset of mimed actions. 

\subsection{The Mimetics dataset}
\label{sub:mimitics}

Mimetics contains short YouTube video clips of mimed human actions that mostly consist in manipulations of, or interactions with certain objects. 
These include sport actions, such as \emph{playing tennis} or \emph{juggling a soccer ball}, daily activities such as \emph{drinking}, personal hygiene, \eg \emph{brushing teeth}, or playing musical instruments including \emph{bass guitar}, \emph{accordion} or \emph{violin}. 
These classes were selected from the action labels of the Kinetics dataset, allowing to evaluate models trained on Kinetics. 
Mimetics contains 713 video clips for a subset of 50 human action classes, \ie, an average of 14.3 clips per class.
As it is hard to find mimed actions on the web, we restrict Mimetics to testing purposes, not for training.
These actions are performed on stage or on the street by mime artists (middle row of Figure~\ref{fig:splash}) but also in everyday life of people, typically during mime games, or captured and shared for fun on social media. 
For instance, the top row of Figure~\ref{fig:splash} shows a video of someone training indoor for \textit{surfing water} or the bottom row shows soccer players mimicking the action \textit{bowling} to celebrate a goal.

Finding videos of mimed action on YouTube is a very difficult task.
The clips for each class were obtained by searching for candidates through the use of key words such as \emph{miming} or \emph{imitating} followed by the desired action, or using query words such as \emph{imaginary} and \emph{invisible} followed by a certain object category. 
We queried these keywords using several different languages but this was not enough to ensure a sufficient number of instances for all the considered action classes. 
To complete the dataset, we also watched hours of videos looking for interesting mimed actions and identifying the clips of interest. 
In comparison, datasets like Kinetics use a semi-automatic process using a frame-level classifier to prune the videos, this was not possible for mimed actions as the classifiers fail.
Some classes had to be dropped due to the lack of videos.
The dataset was built making sure that a human observer was able to recognize the mimed actions. 
The  videos have variable resolutions and frame rates and have been manually trimmed between 1 and 10 seconds, following the Kinetics dataset. 
The URLs of the original YouTube videos and the temporal intervals of the video clips have been shared to spur further research on this topic.
The detailed list of classes with the number of videos per class is available in Appendix~\ref{app:results}.

\subsection{Experimental results}

\begin{figure*}
 \includegraphics[width=\linewidth]{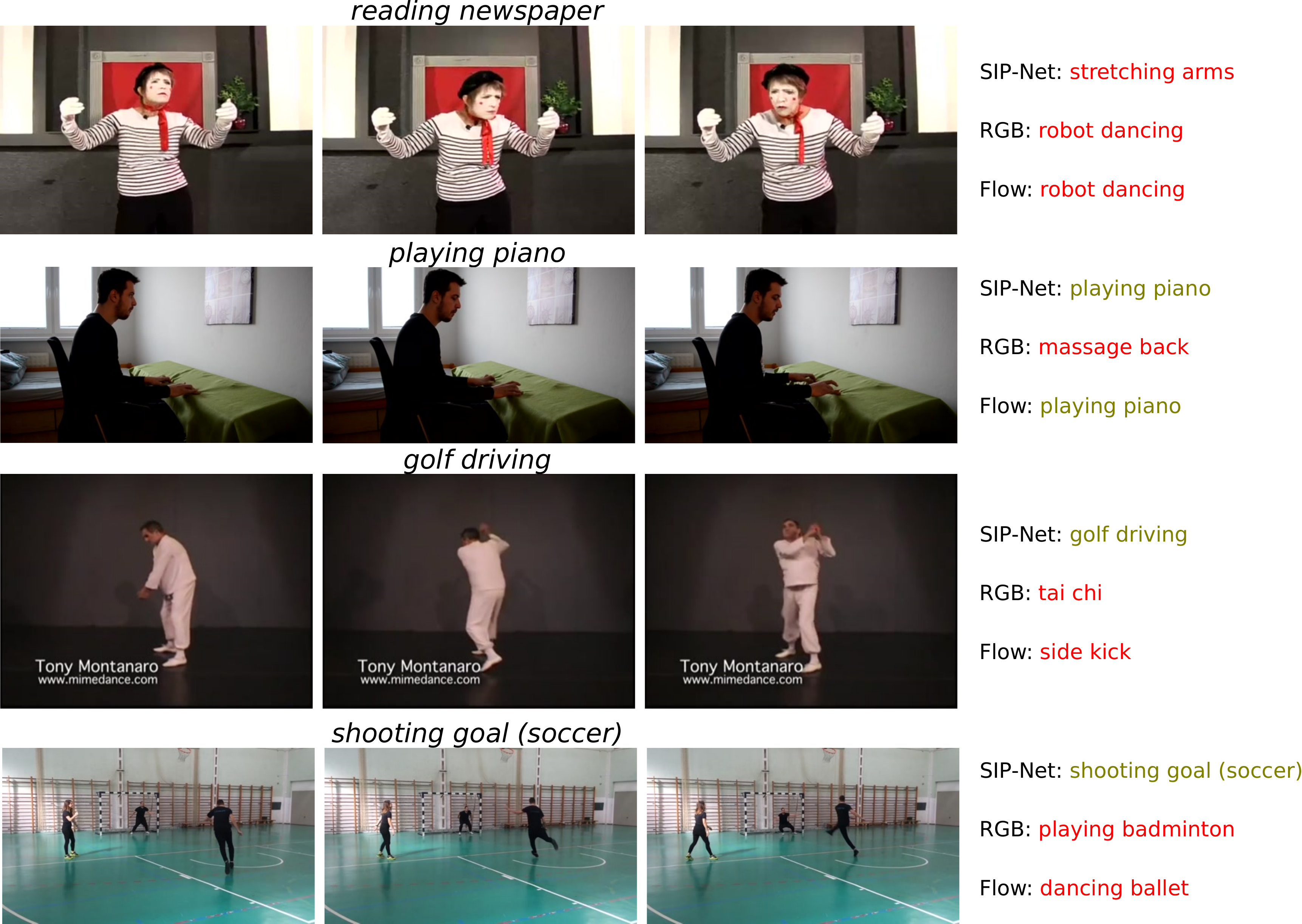}
 \caption{Four video examples from Mimetics with the highest scored class for the SIP-Net, RGB 3D CNN and Flow 3D CNN}
 \label{fig:results}
\end{figure*}

We compare several approaches on the Mimetics dataset: our three pose-based baselines, a state-of-the-art 3D CNN method on RGB input or Flow input as well as their late fusion, in addition to STGCN~\citep{STGCN} with OpenPose. For optical flow input, we use the TV-L1 algorithm~\citep{TVL1}.
All methods were trained on the 400 classes of Kinetics.
We then run them on the videos from the Mimetics dataset, and report top-1, top-5 accuracies as well as the mean average-precision (mAP).
As each video has a single label, average-precision computes for each class the inverse of the rank of the ground-truth label, averaged over all videos of this class.
Overall performances are reported in Table~\ref{tab:mimetics}. We refer to Appendix~\ref{app:results} for per-class results.
Figure~\ref{fig:results} shows some qualitative examples.

\begin{table}
 \caption{Mean top-k accuracies and mean average-precision (in \%) on the Mimetics dataset when training on Kinetics}
 \centering
 \begin{tabular}{l|rr|r}
   & top-1 & top-5 & mAP\\
\hline
RGB (3D-ResNeXt-101)                     &    8.6 &   20.1 &    15.6 \\
Flow (3D-ResNeXt-101)                    &   11.8 &   29.6 &    21.1 \\
RGB+Flow (late fusion)                   &   10.5 &   26.9 &    19.1 \\
STGCN (OpenPose)                         &   12.6 &   27.4 &    20.7 \\
\hline
STGCN2D                                  &    9.0 &   20.5 &    15.4 \\
STGCN3D                                  &    5.8 &   13.8 &    11.3 \\
SIP-Net                                  &   {\bf14.2} &  {\bf32.0} &  {\bf22.7} \\
\hline
 \end{tabular}
 \label{tab:mimetics}
\end{table}

We first observe that the performance is relatively low for all methods, below 15\% top-1 accuracy and 25\% mAP, showing that the recognition of mimed actions is challenging.
In fact, all methods completely fail for a certain number of actions including \textit{climbing a rope}, \textit{reading newspaper}, \textit{eating cake} or, more surprisingly, \textit{sweeping floor}.
One reason for this overall low accuracy is that some Kinetics actions are fine-grained (\eg different classes correspond to \textit{eating} various types of food) and are hard to distinguish, especially when mimed.
However, even when considering higher-level `superclasses', the performance remains quite low as shown in Appendix~\ref{app:superclass}.
Another difficulty is that mimed actions tend to be exaggerated, some in a comical way but also to virtually represent the object. This is  particularly true when they are performed by mime artists. For instance, in the  \textit{reading newspaper} sequence of Figure~\ref{fig:results}, the artist exaggerates the movements of the head to make people understand that he/she is reading.  These gestures are consequently not aligned with real performances of the actions as observed in the training videos. Interestingly, a person who has never seen a mime before is still capable of understanding what is happening and so should an intelligent system.
We manually label a flag for each video whether the actor is a mime artist or not, and show the global top-1 accuracy in Table~\ref{tab:top1vid}.
For all approaches, the performance is significantly lower on videos where actions are performed by mime artists compared to standard people.

The best overall performance is achieved by SIP-Net which consists of a temporal convolution applied on pose features, 
reaching 14.2\% top-1 accuracy and a mAP of 22.7\%. 
Some failure cases occur when several people are present in the scene. 
The tubes can erroneously mix several individuals or other persons (\eg spectators) sometimes obtain higher scores than the one miming the action of interest.

\begin{figure}
 \resizebox{\linewidth}{!}{
 \includegraphics[height=3cm]{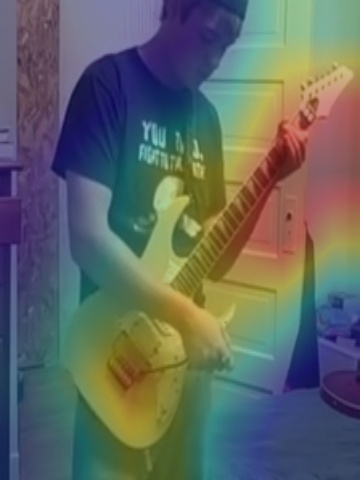}
 \includegraphics[height=3cm]{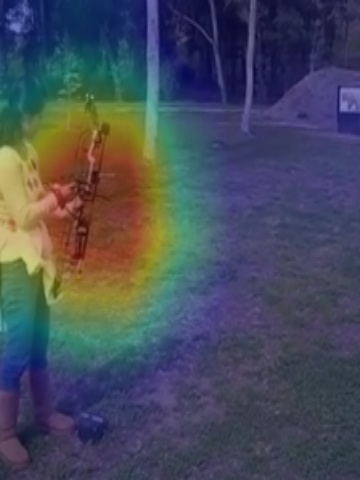}
 \includegraphics[height=3cm]{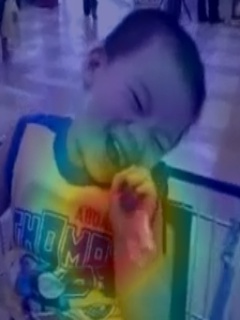}
 \includegraphics[height=3cm]{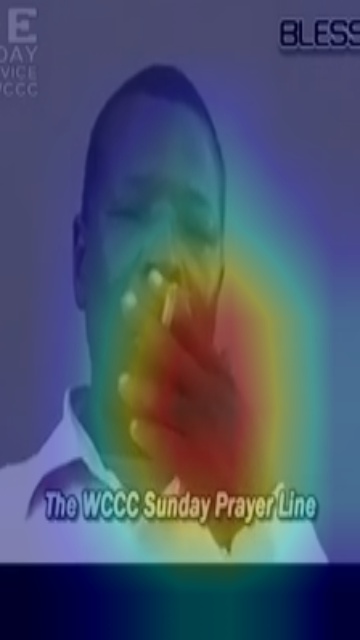}
 }
 \caption{Class Activation Maps for the central frame of the central clip of several Kinetics validation video. The left two examples show classes with a relevant and large objects (\textit{playing guitar} and \textit{archery}) while the two right examples involve smaller objects (\textit{brushing teeth} and \textit{smoking}).}
 \label{fig:CAM}
\end{figure}

In comparison, state-of-the-art 3D CNN model trained on RGB clips performs more poorly, with 8.6\% mean top-1 accuracy and 15.6 mAP.
For some classes such as \textit{archery}, \textit{playing accordion}, \textit{playing bass guitar}, \textit{playing trumpet}, this state-of-the-art RGB model obtains 0\% while SIP-Net performs decently. 
One key reason for that is the bias learned by the model: it focuses on the objects being manipulated or the scenes where the video is captured more than on the performed actions.
For instance, in the second row of Figure~\ref{fig:results}, someone mimics \emph{playing piano} on a console table covered with a tablecloth, which looks like a massage table.
As a consequence, the RGB model predicts the action \textit{massage back} without considering what the person is really doing.
To verify this bias towards objects and scenes, we show in Figure~\ref{fig:CAM} (left) the Class Activation Maps (CAM)~\citep{CAM} for two examples of classes that involve large objects, namely for the class \textit{playing guitar} and \textit{archery}: in both cases, the network mainly leverages areas covered by the large objects involved in the action more than the actor.
To further verify the bias towards objects and scenes, we manually label for each video if there is any relevant object or not, and if the scene is relevant for the action.
We report the global top-1 accuracy (as some classes have no video or just a few, global accuracy is better suited than mean per-class accuracy) in Table~\ref{tab:top1vid} for the subset of videos where there is no relevant object, where the scene is not relevant or both. On these videos, the performance of the state-of-the-art RGB 3D CNN  significantly drops while the SIP-Net baseline is more robust.
RGB 3D CNN still performs better than SIP-Net on classes such as \textit{brushing teeth}, \textit{catching or throwing baseball},  or \textit{juggling balls}.
This corresponds to classes in which the object is barely visible in most training videos, either too small (\eg cigarette for \emph{smoking}) or mostly occluded by hands (baseball ball, toothbrush, hair brush). 
In such cases, 3D CNN model focuses on face and hands (for \textit{brushing teeth}, \textit{smoking}) or on the body (\textit{throwing baseball}) and therefore performs reasonably well on these mimed actions. 
To confirm this, we computed  the CAM~\citep{CAM} for classes with small objects,  shown in Figure~\ref{fig:CAM} (right), and observe that the area contributing the most covers the hands and/or the face.
To further verify this, we manually annotate for each of the 50 classes of Mimetics whether there is an object being manipulated or not, and if it is small or large.
We report the global top-1 accuracy in Table~\ref{tab:top1cls}.
RGB performs better than SIP-Net on actions with no object or with small objects, while SIP-Net clearly outperforms RGB in case of large objects.

One hypothesis could be that a better pretraining of 3D CNNs models based on RGB may lead to higher results. To verify that, we evaluated the model from~\cite{ig65m} that uses weakly-supervised pretraining on 65 million social-media videos based on hashtags, before finetuning on Kinetics. This model is based on a (2D+1)-convolutional ResNet-34 architecture and obtains 11.3\% top-accuracy, 22.9\% top-5 accuracy and 17.7\% mAP.
This is about 2-3\% above the 3D-ResNeXt-101 RGB baseline but remains quite low, and in particular worse than models based on Flow or SIP-Net.

\begin{table}
 \caption{Global top-1 accuracy (in \%) on various subsets of Mimetics for models trained on Kinetics}
 \centering
 \resizebox{\linewidth}{!}{
  \begin{tabular}{lc|rr|r}
                               & (\#vid.) & RGB & Flow & SIP-Net \\
 \hline
    all videos                 & (713) & 8.4 & 11.5 & {\bf14.3} \\
 \hline
    mime artist                & (203) & 4.9 & {\bf6.4} & 5.4 \\
 not a mime artist             & (510) & 9.8 & 13.5 & {\bf17.8} \\
 \hline
 no object is relevant         & (644) & 6.8 & 9.8 & {\bf13.4} \\
 scene is not relevant         & (644) & 6.4 & 9.8 & {\bf13.5} \\
 no object is relevant, and scene is not relevant & (584) & 4.5 & 8.0 & {\bf12.7} \\
  \hline
  \end{tabular}
 }
 \label{tab:top1vid}
\end{table}

\begin{table}
 \caption{Global top-1 accuracy (in \%) for classes with no/small object and with large objects}
 \centering
 \resizebox{\linewidth}{!}{
  \begin{tabular}{lcc|rr|r}
                         & \#cls. & (\#vid.) & RGB & Flow & SIP-Net \\
 \hline
    all classes          & 50     & (713) & 8.4 & 11.5 & {\bf14.3} \\
 \hline
    no/small object      & 19     & (268) & 11.2 & {\bf12.3} & 9.0 \\
    large object        & 31     & (445) &  6.7 & 11.0 & {\bf17.5} \\
  \hline
  \end{tabular}
 }
 \label{tab:top1cls}
\end{table}

We then also evaluate a similar 3D CNN that takes as input optical flow clips instead of RGB clips.
The overall performance is higher than RGB, with 11.8\% top-1 accuracy and 21.1\% mAP.
This suggests that this flow model learns less biases than RGB, because it does not see the appearance of the scenes and objects.
For instance, \emph{playing piano} is correctly predicted in the example of the second row of Figure~\ref{fig:results}, because a piano and a covered table roughly look the same from an optical flow point of view.
\cite{sevilla2018integration} suggest that flow may still capture global shape of the actor or objects.
This explains why, as RGB, flow model performs better on classes with small or no object compared to classes involving larger objects, see Table~\ref{tab:top1cls}: when the subject is manipulating small objects, the network is not able to capture these details and it focuses on bigger structure like the person, thus generalizing better to out-of-context actions.
We evaluated in Table~\ref{tab:mimetics} a late fusion of RGB and Flow, \ie, a two-stream model~\citep{TwoStream}, and observe a small decrease of performance as both models tend to perform well on the same classes and videos.

Next, we also benchmark other pose-based approaches.
Our two baselines based on explicit 2D or 3D poses perform quite poorly, comparably to their respective performance on the Kinetics dataset.
This can be explained by the difficulty to extract accurate body keypoint coordinates for videos in-the-wild with abrupt camera and actor motion, blur, and occlusions. 
In particular, the low performance on Kinetics itself suggests this occurs also in the training set, leading to a poor model.
We also compare to STGCN~\citep{STGCN} that uses OpenPose to estimate the pose, \ie, with more keypoint on the head than LCR-Net++.
The performance is higher with 12.6\% top-1 accuracy but remains lower than the SIP-Net baseline that does not explicitly compute poses but transfers the learned pose features to action recognition.

To explain the relatively poor performance of all methods, we argued that Kinetics classes might be too fine-grained and too difficult to distinguish when mimed.
This is illustrated by the significantly higher top-5 accuracy (32.0\%) than top-1 accuracy (14.2\%), see Table~\ref{tab:mimetics}.
To further verify this statement, we trained a SIP-Net model on the Kinetics training videos from the 50 classes of Mimetics and report the results in Table~\ref{tab:mimetics50}.
Top-1 accuracy increases to more than 25\% and top-5 accuracy to more than 50\%.

\begin{table} 
 \caption{Mean top-k accuracies and mAP (in \%) of SIP-Net on the Mimetics dataset when training on the full Kinetics training set, or on the subset of classes from Mimetics}
 \centering
 \resizebox{\linewidth}{!}{
 \begin{tabular}{l|rr|r}
training set  & top-1 & top-5 & mAP\\
\hline
Kinetics (400 classes)                 &     14.2 &       32.0  &       22.7 \\
Kinetics subset (50 classes)           & {\bf25.1} &  {\bf51.4} &   {\bf38.3} \\
\hline
 \end{tabular}
 }
 \label{tab:mimetics50}
\end{table}

\section{Conclusion}

In this paper, we have highlighted the context biases of existing action recognition datasets and 3D CNN models.
To benchmark performances on out-of-context actions, we have introduced the Mimetics dataset.
Our experiments show that models leveraging body language via human pose are less prone to the context biases.
Applying a shallow neural network such as a single convolution over features transferred from human poses performs surprisingly well compared to 3D action recognition applied in-the-wild.
Our analysis shows that using a sparse set of keypoints might not be sufficient to distinguish some fine-grained actions.
Using a more complete representation of human poses including full-body, hands, and face dense pose information, as predicted by recent works in human pose/shape estimation could significantly increase the performance.

We think that our new Mimetics benchmark will allow to better understand what action recognition models learn and is a step towards designing more intelligent systems. We hope it will stimulate research into the particular challenges of out-of-context action recognition. Estimating the performance of future action recognition methods on
our Mimetics dataset could help bringing additional analysis on their similarity to human performance but it could also help evaluate their capability to detect/ignore mimes. 
Ideally, an action recognition system should solve both the problems of recognizing a human action, and identifying whether it is mimicked or not (fake or real). 
Our work also allows to make a step toward this goal by showing how much state-of-the-art action recognition systems can be fooled by mimes. This particularly occurs when the context is partial, see the second case in Figure~\ref{fig:results} classified as `massage back'. The mistakes made by these models in such scenarios are harmful for their real-life deployments.

\bibliographystyle{spbasic}      % basic style, author-year citations
\bibliography{biblio}   % name your BibTeX data base

\clearpage

\appendix

\section{Extended experiments on existing datasets}
\label{app:xp}

In this section, we provide more analysis about the performance of the three pose-based baselines on existing action recognition datasets.
We first perform a parametric study of SIP-Net in Section~\ref{sub:sipnet}.
We then use the various levels of ground-truth (see Table 3 of the main paper) to study the impact of using ground-truth or extracted tubes and poses (Section~\ref{sub:baselines}).

\noindent \textbf{Tubes.}
For datasets with ground-truth 2D poses, we compare the performance when using ground-truth tubes (GT Tubes) obtained from GT 2D poses, or estimated tubes (LCR Tubes) built from estimated 2D poses, see Section 4.1 of the main paper.
In the latter case, tubes are labeled positive if the spatio-temporal IoU with a GT tube is over $0.5$, and negative otherwise. 
When there is no tube annotation, we assume that all tubes are labeled with the video class label.
Note that in some videos, no tube is extracted, in which case the videos are ignored when training, and considered as wrongly classified for test videos.
In particular, this happens when only the head is visible, as well as for many clips with first person viewpoint, where only one hand or the main manipulated object is visible.
We obtain no tube for 0.1\% of the videos on PennAction, 2.5\% on JHMDB, 2.7\% on HMDB51, 6.7\% on UCF101 and 15.3\% on Kinetics.

\subsection{SIP-Net baseline}
\label{sub:sipnet}
 
\begin{figure}
\centering
\includegraphics[width=0.49\linewidth]{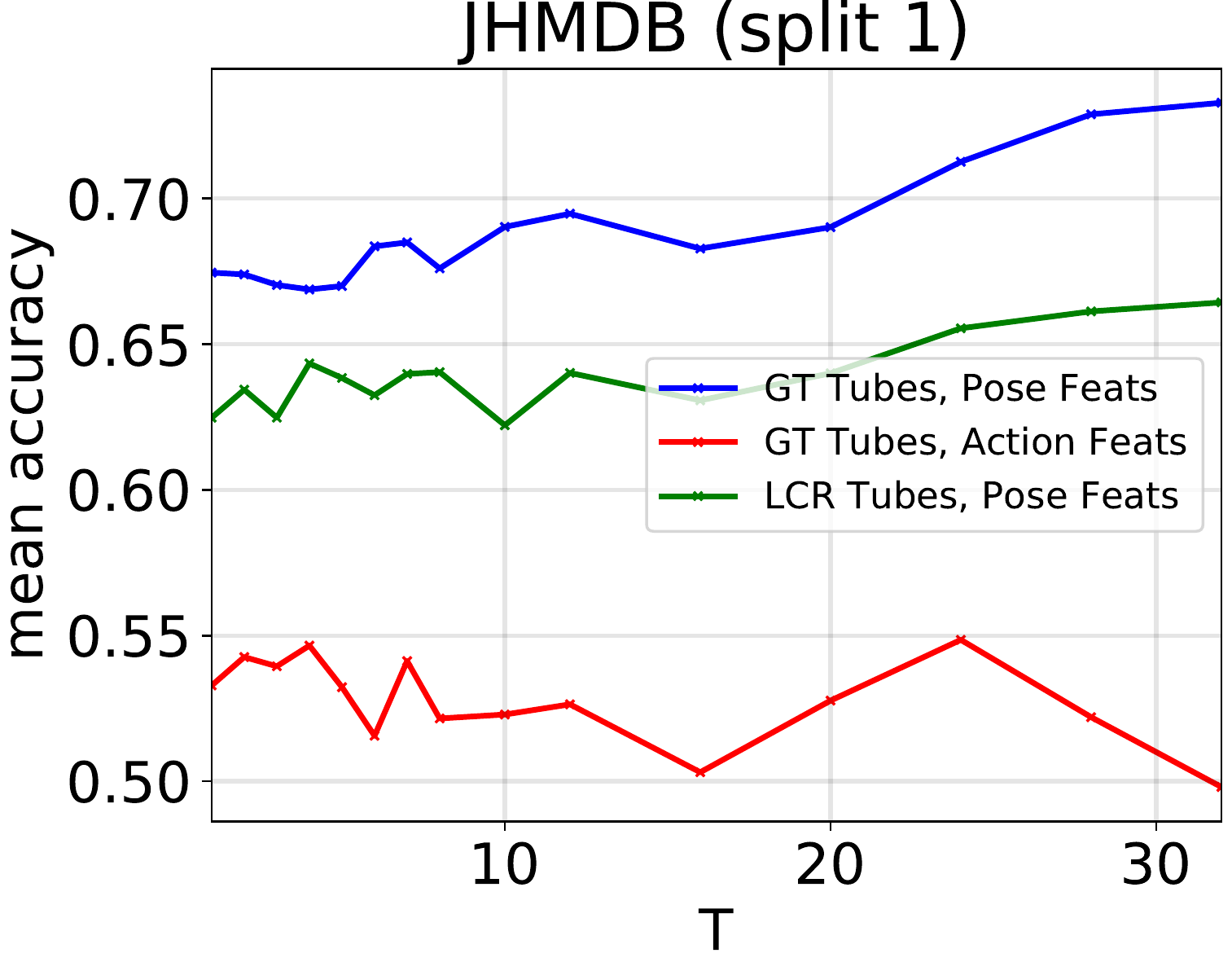} \hfill
\includegraphics[width=0.49\linewidth]{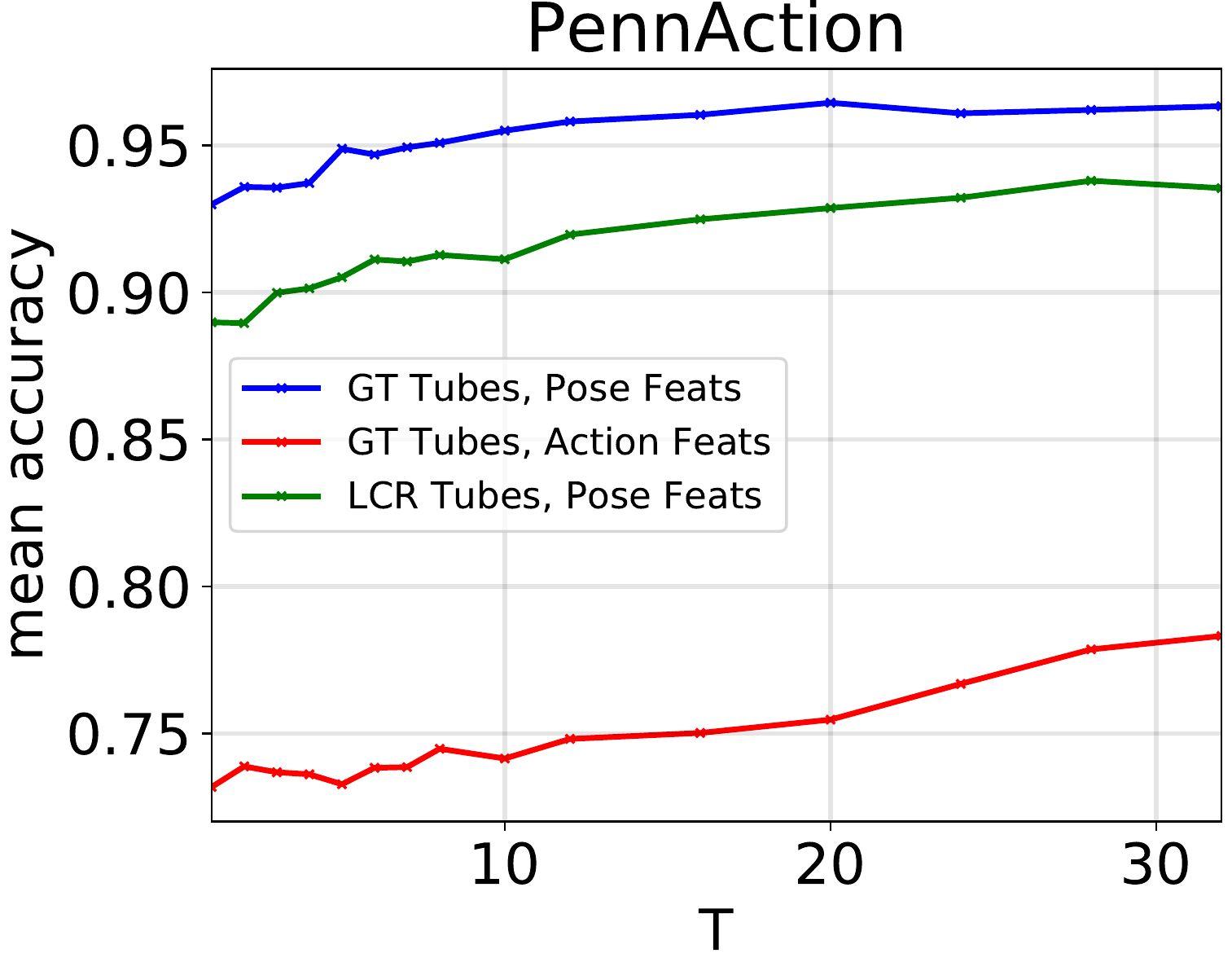} \\
\includegraphics[width=0.49\linewidth]{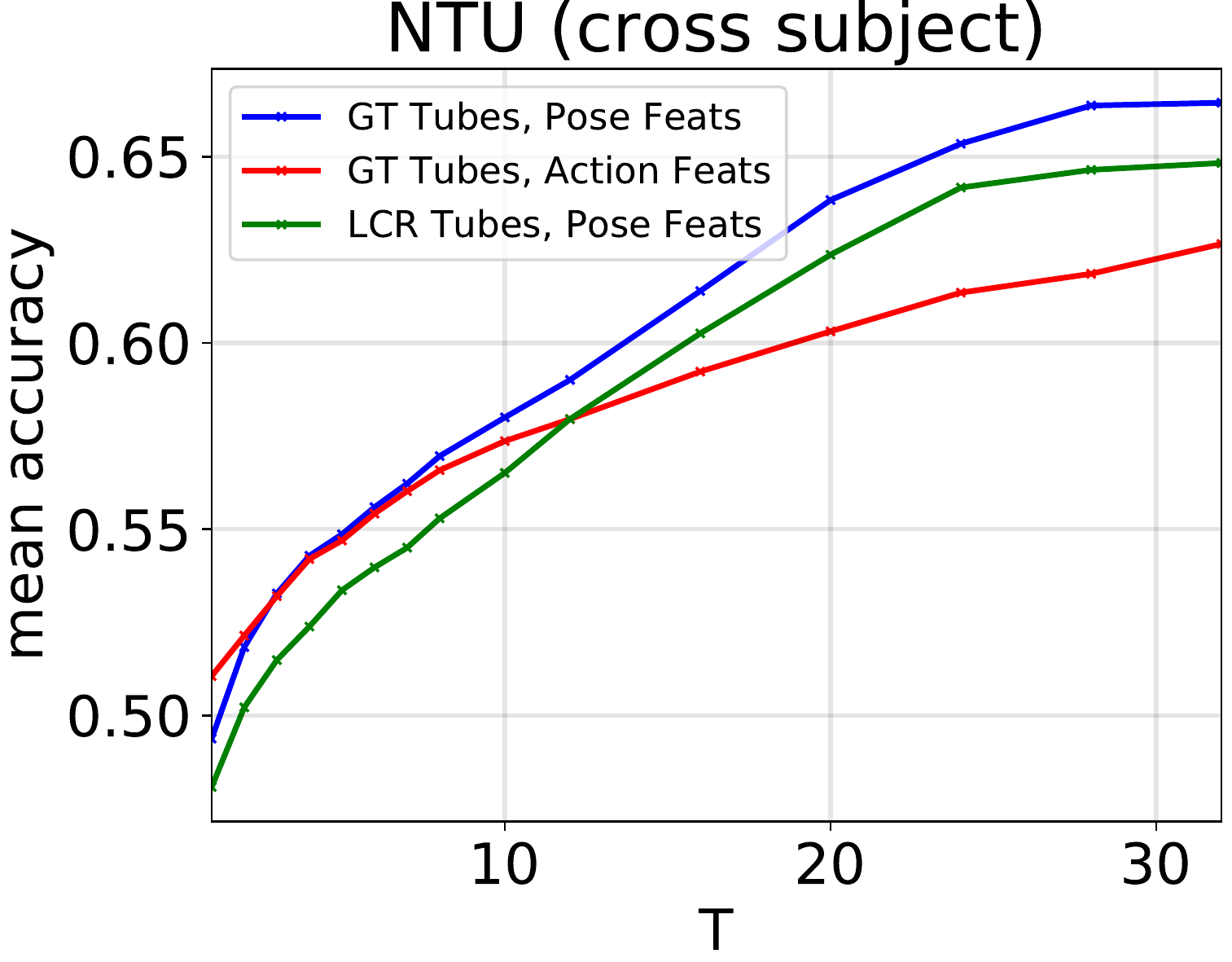} \hfill
\includegraphics[width=0.49\linewidth]{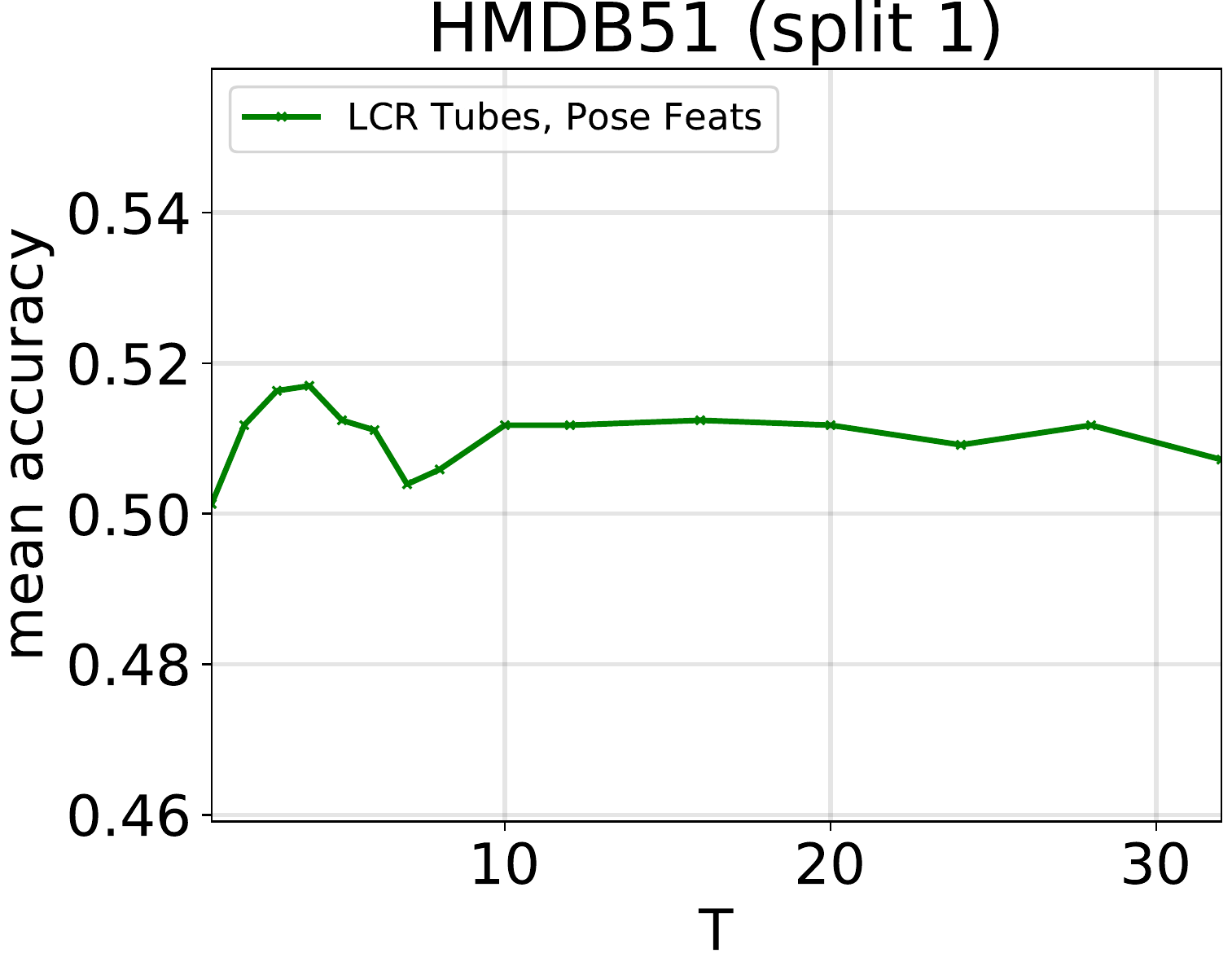} \\
\includegraphics[width=0.49\linewidth]{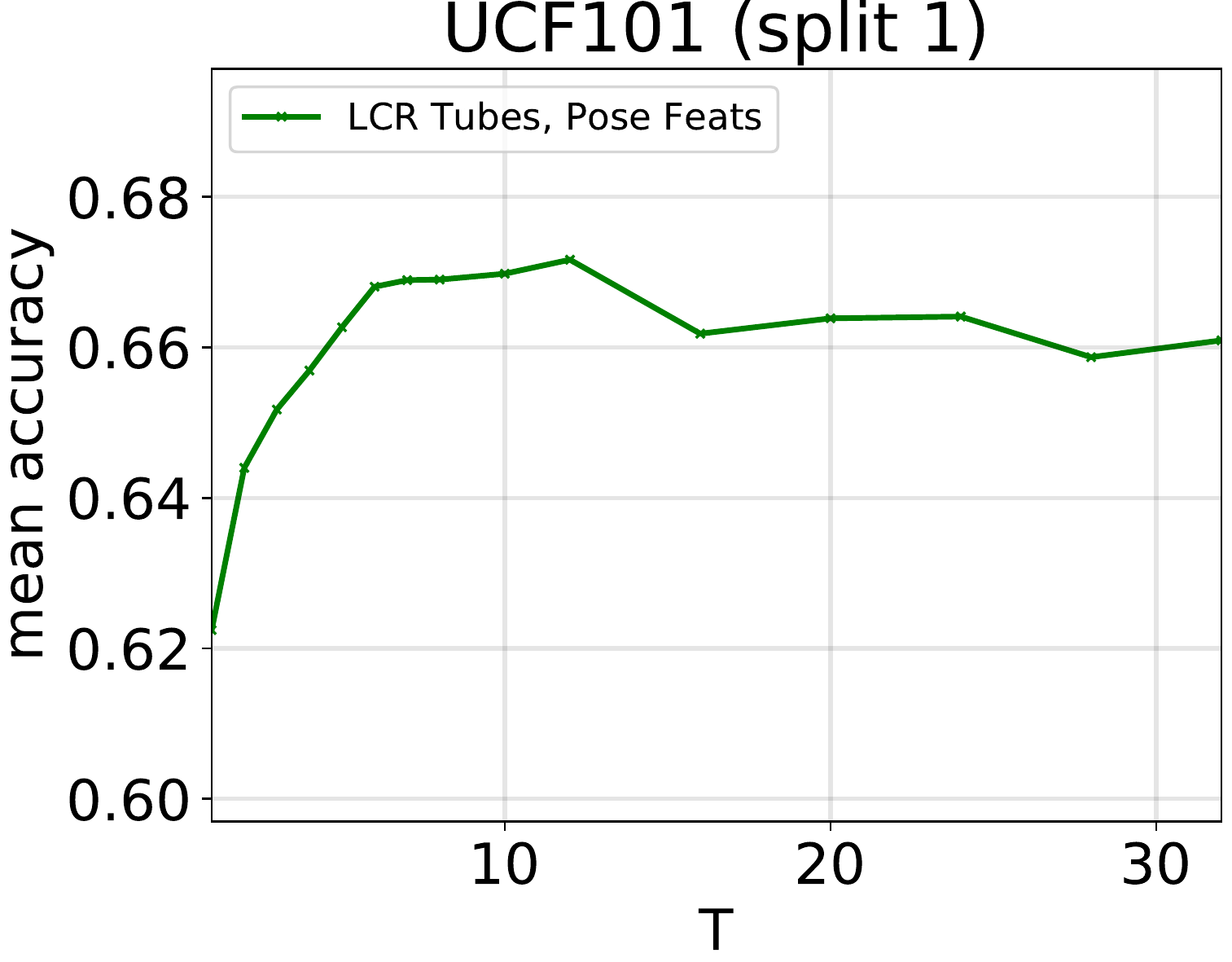} \hfill
\includegraphics[width=0.49\linewidth]{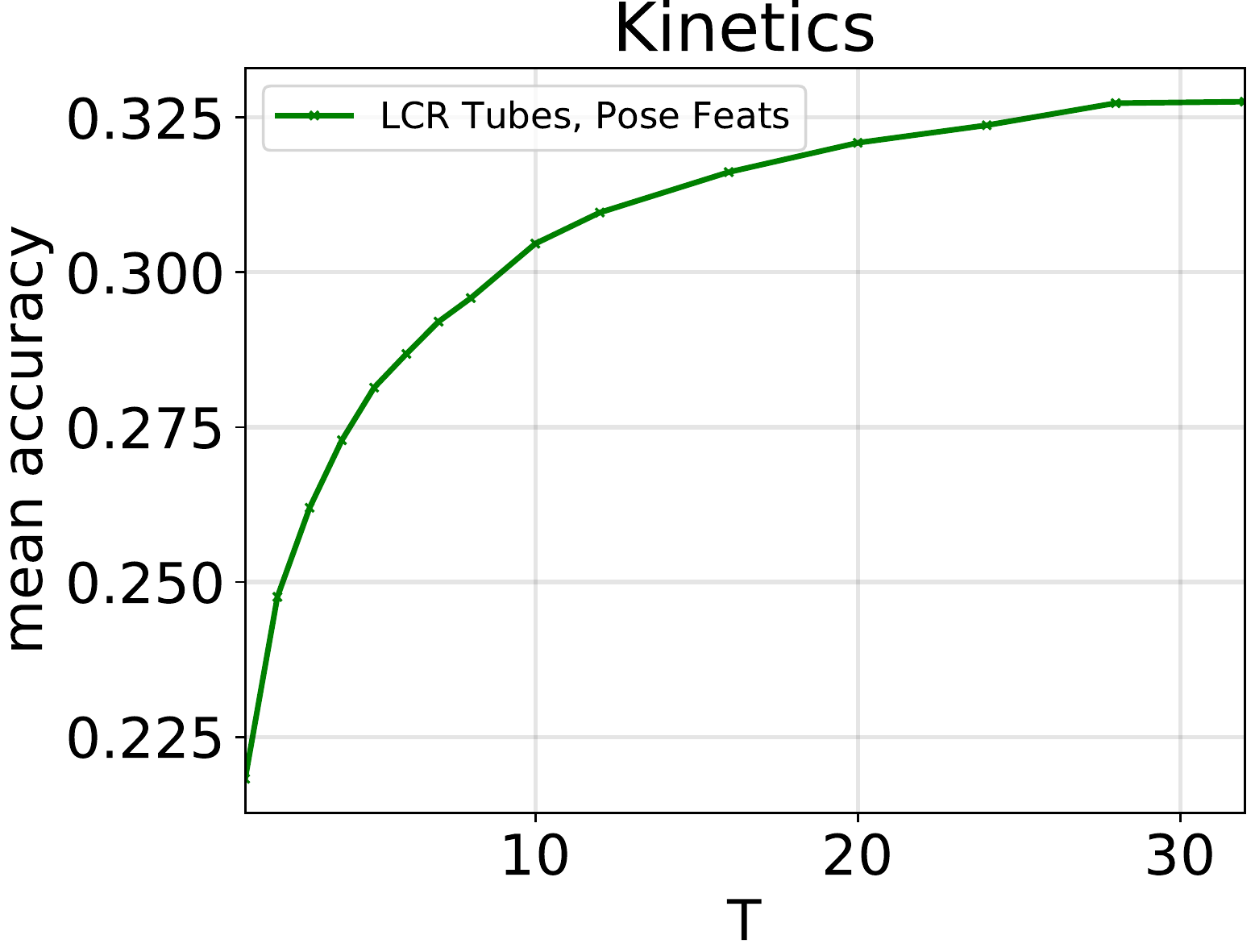} 
\caption{Mean-accuracy of SIP-Net for varying $T$ on all datasets, for different tubes (GT or LCR) and features (Pose or Action).}
\label{fig:T}
\end{figure}

We first present the results for the SIP-Net baseline with GT tubes (blue curve `GT tubes, Pose Feats') and LCR tubes (green curve `LCR Tubes, Pose Feats') on all datasets for varying clip length $T$, see Figure~\ref{fig:T}.
Overall, a larger clip size $T$ leads to a higher classification accuracy.
This is in particular the case for datasets with longer videos such as NTU and Kinetics.
This holds both when using GT tubes (blue curve) and LCR tubes (green curve).
We keep $T$=32 in the remaining of this paper.
 
\begin{figure}
\centering
\includegraphics[width=0.9\linewidth]{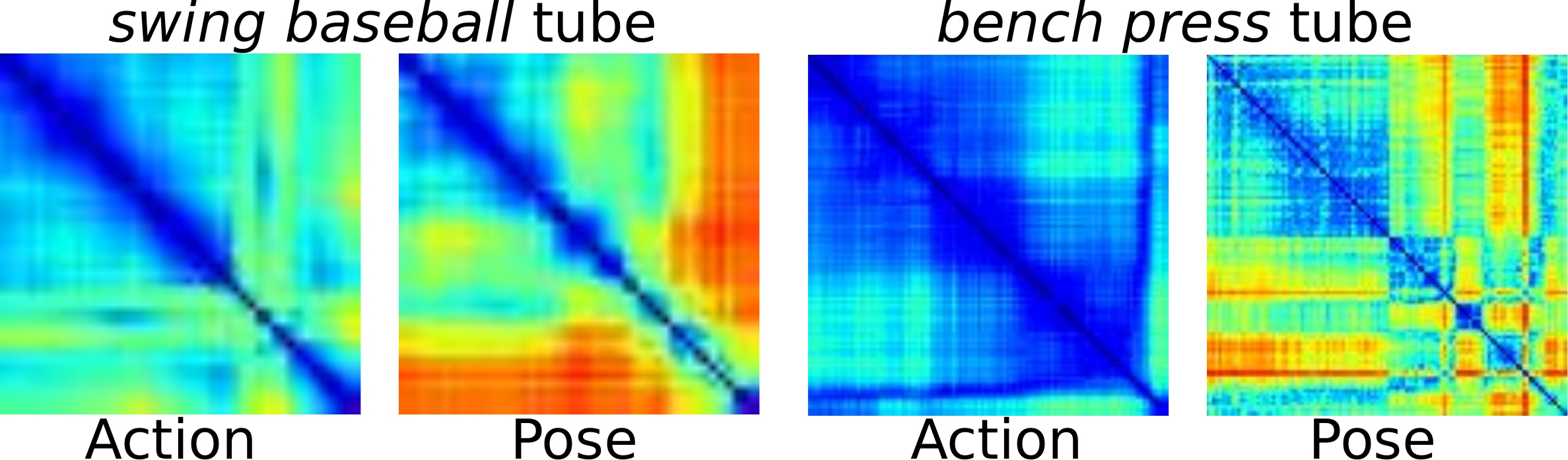}
\caption{Feature correlation for two different videos of PennAction (action \textit{swing baseball} on the left and \textit{bench press} on the right). For each sequence, we show the distances between features along the tube when using Faster R-CNN action or LCR pose features. Blue/red color indicates low/high distances and therefore high/low correlation. Implicit pose features clearly show more variation inside a tube.}
\label{fig:corr}
\end{figure}

\begin{figure}
\centering
\includegraphics[width=0.49\linewidth]{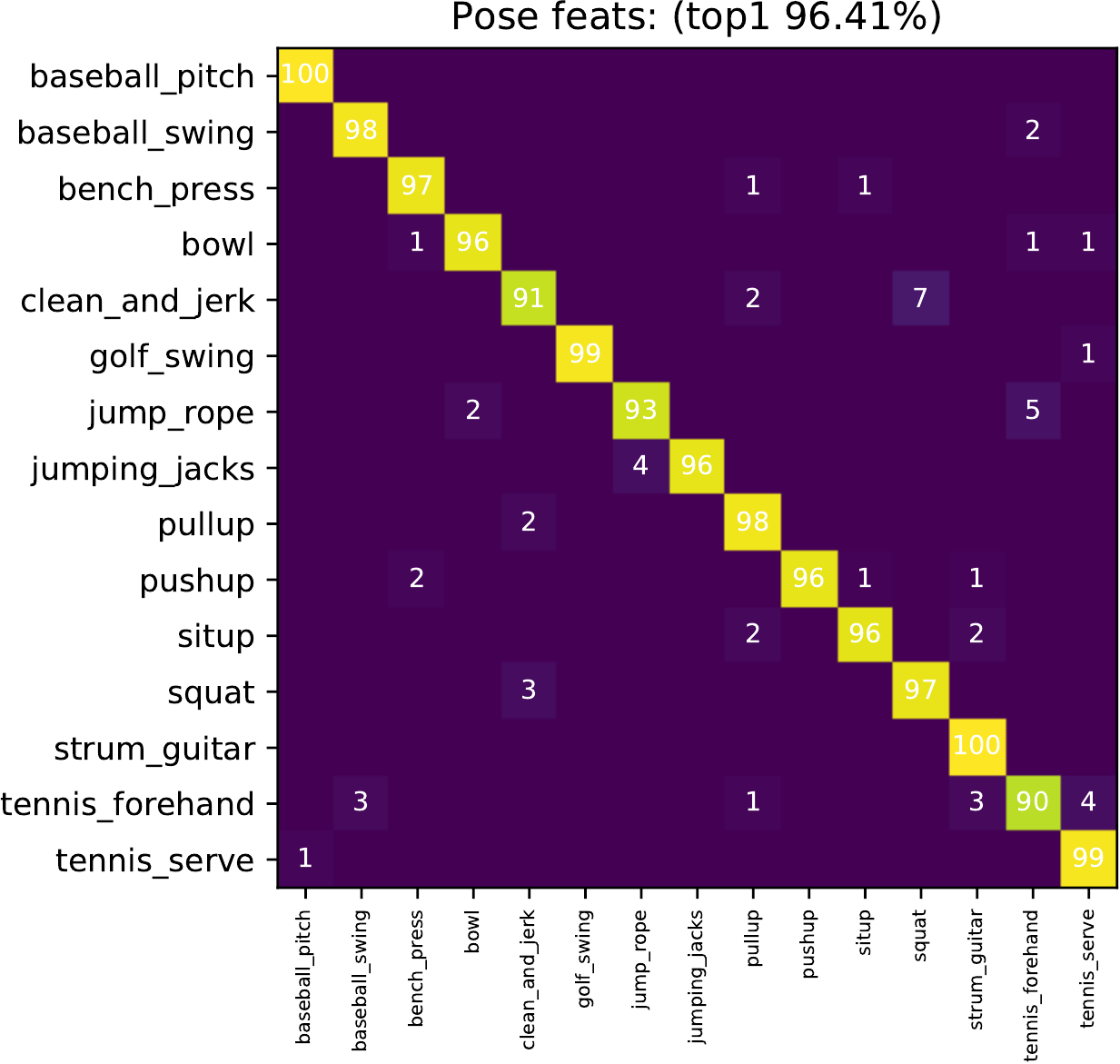}
\hfill
\includegraphics[width=0.49\linewidth]{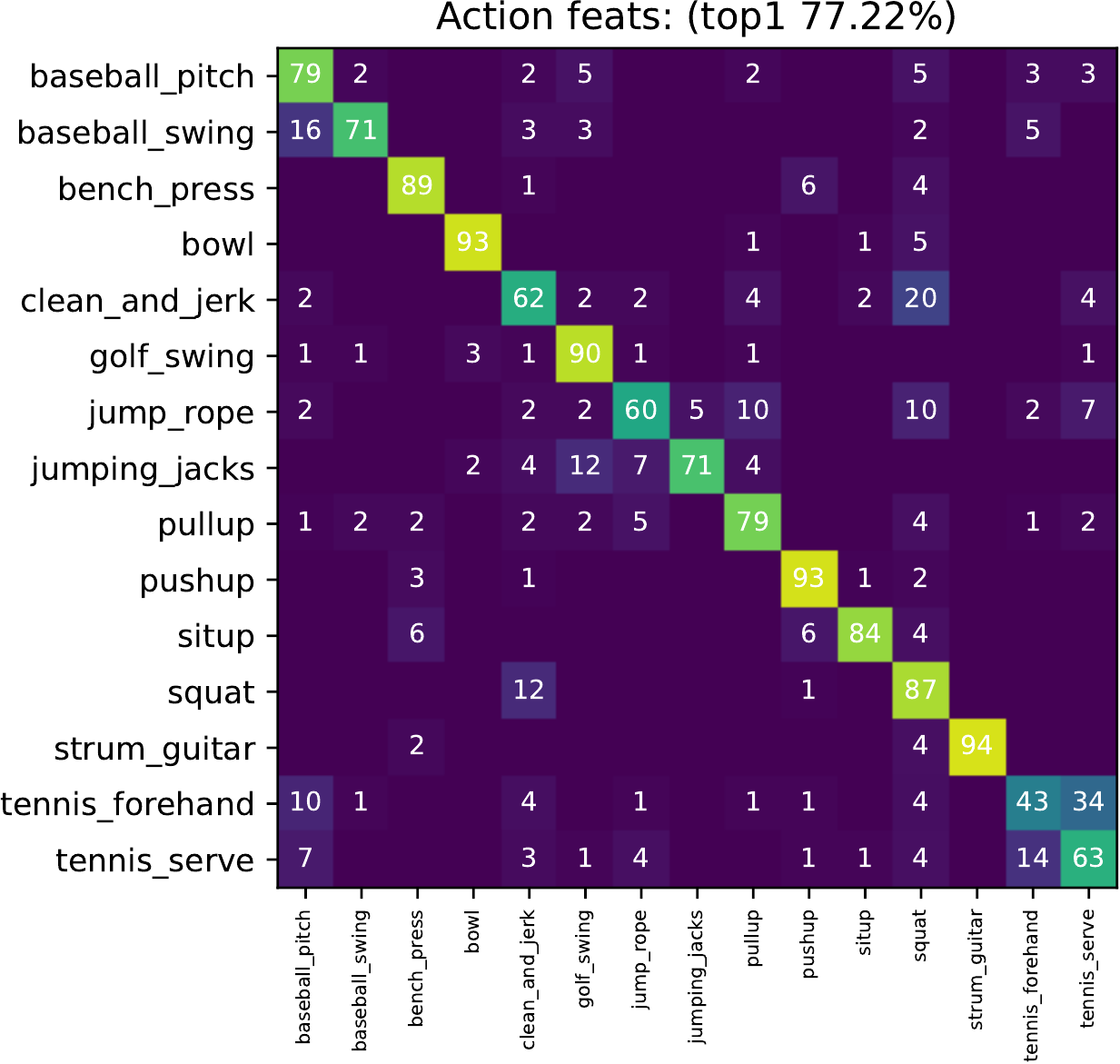}
\caption{Confusion matrices on PennAction when using action features (left) and pose features (right) in SIP-Net.}
\label{fig:confusion}
\end{figure}

Next, we measure the impact of applying transfer learning from the pose domain to action recognition.
To this end, we compare the temporal convolution on LCR pose features (blue curve, `Pose Feats'), to features extracted from a Faster R-CNN model with ResNet50 backbone trained to classify actions (red curve, `Action Feats').
This latter method is not supposed to be state-of-the-art in action recognition, but it allows to fairly compare the pose features to action features, keeping the network architecture exactly the same, simply changing the learned weights.
Note that such a frame-level action detector has been used in the spatio-temporal action detection literature~\citep{saha2016deep,weinzaepfel2015learning}, before the rise of 3D CNNs.
Results in Figure~\ref{fig:T} show a clear drop of accuracy when using action features instead of pose features:
about 20\% on JHMDB-1 and PennAction, and around 5\% on NTU for $T$=32.
Interestingly,  this holds for T=1 on HMDB-1 and PennAction, \ie, without temporal integration, showing that `Pose feats' are more powerful.
To better understand why using pose features considerably increases performance compared to action features, we visualize the distances between features inside tubes in Figure~\ref{fig:corr}.
When training a per-frame detector specifically for actions, most features of a given tube are correlated. It is therefore hard to leverage temporal information from them.
In contrast, LCR-Net++ pose features considerably change over time, as does the pose, deriving greater benefit from temporal integration.
Figure~\ref{fig:confusion} shows confusion matrices on PennAction when using `Pose feats' (left) \vs `Action feats' (right). 
With `Action feats', confusions happen between the two \textit{tennis} or the two \textit{baseball} actions, while this is disambiguated with `Pose feats'.

\subsection{Comparison between baselines}
\label{sub:baselines}

We  compare the performance of the  baselines using GT and LCR tubes, on the JHMDB-1, PennAction and NTU datasets in Table~\ref{tab:GT}.
On JHMDB-1 and PennAction, despite being a much simpler architecture, the SIP-Net baseline outperforms the methods based on explicit 2D-3D pose representations, both with GT and LCR tubes.
Estimated 3D pose sequences are usually noisy and may lack temporal consistency. 
We also observe that the STGCN3D approach significantly outperforms its 2D counterpart (STGCN2D), confirming that 2D poses contain less discriminative and more ambiguous information.

On the NTU dataset, the 3D pose baseline obtains 74.8\% accuracy when using GT tubes and estimated poses (STGCN3D on GT Tubes), compared to 81.5\% reported in~\citep{STGCN} when using ground-truth 3D poses.
This gap of 7\%  in a constrained environment is likely to increase for videos captured in the wild. 
The performance of the features-based baseline (SIP-Net) is lower, 66.4\% on GT tubes, suggesting than SIP-Net performs better only in unconstrained scenarios.

\begin{table}
\centering
\caption{Mean accuracies (in \%) for our three baselines on datasets with GT tubes, or when using LCR tubes}
\resizebox{\linewidth}{!}{
\begin{tabular}{l|c|ccc}
 \multicolumn{1}{l}{Method} & Tubes & JHMDB-1 & PennAction & NTU (cs) \\
  \hline 
  STGCN2D           &  \multirow{3}{*}{GT}  &  34.9 &  90.8 &  66.3 \\
  STGCN3D           &                       &  57.9 &  94.7 & {\bf74.8} \\
  SIP-Net           &                       & {\bf 73.3} & {\bf  96.3 }&  66.4 \\
  \hline
  STGCN2D           & \multirow{3}{*}{LCR}  &  23.2 &  85.5 &  69.4 \\
  STGCN3D           &                       &  53.1 &  89.2 &  {\bf 75.0} \\
  SIP-Net           &                       &  {\bf 66.4 }&  {\bf 93.5 }&  64.8 \\
  \hline
  \multicolumn{2}{c|}{STGCN3D (GT 3D poses)}  &    -    &    -    &  81.5 \\
  \hline
\end{tabular}
}
\label{tab:GT}
\end{table}

\section{Per-class results on Mimetics}
\label{app:results}

In Table~\ref{tab:results}, we present for each class the top-1 accuracy and the AP of the different methods.
For the top-1 accuracy metric, SIP-Net obtains the best performance for 19 out of 50 classes, with a mean accuracy of 14.2\%.
The RGB 3D CNN baseline obtains the highest AP for 8 classes, which often correspond to classes 
in which manipulated objects are small, making the network less bias towards context (\eg the ball for the action \textit{catching of throwing baseball}).
Table~\ref{tab:results} also highlights that the recognition of mimed actions is a very challenging and open task, 
as none of the videos are correctly classified (\ie 0\% top-1 accuracy) by the 5 baselines for 5 out of the 50 classes.

\begin{table*}
 \centering
 \caption{Per-class results on the Kinetics for the RGB and Flow 3D CNN baselines, for STGCN~\citep{STGCN} (2D with OpenPose), as well as our three baselines (STGCN2D, STGCN3D, SIP-Net).
 In each column, the first number is the top-1 accuracy per class (in \%), and the number in parenthesis is the AP (in \%).}
 \resizebox{\linewidth}{!}{
 \begin{tabular}{lr||rr|rr||rr||rr|rr|rr}
 class                                             & \#vid & \multicolumn{2}{c|}{RGB} & \multicolumn{2}{c||}{Flow} & \multicolumn{2}{c||}{STGCN} & \multicolumn{2}{c|}{STGCN2D} & \multicolumn{2}{c|}{STGCN3D} & \multicolumn{2}{c}{SIP-Net} \\
\hline                                                                                                                                                                                                         
archery & 19 & 0.0 & (3.4)& 0.0 & (2.6)& 5.3 & (11.3)& 0.0 & (9.5)& 0.0 & (10.7)& \textbf{36.8} & \textbf{(42.0)} \\                                                                                         
bowling & 13 & \textbf{15.4} & (16.8)& \textbf{15.4} & \textbf{(21.1)}& 0.0 & (2.4)& 0.0 & (4.9)& 0.0 & (3.4)& 7.7 & (15.9) \\                                                                               
brushing hair & 20 & 15.0 & (23.6)& \textbf{25.0} & \textbf{(39.6)}& 0.0 & (8.6)& 0.0 & (2.7)& 0.0 & (1.0)& 0.0 & (7.5) \\                                                                                   
brushing teeth & 15 & 40.0 & (45.4)& \textbf{53.3} & \textbf{(62.8)}& 13.3 & (24.4)& 0.0 & (1.7)& 0.0 & (1.9)& 6.7 & (25.3) \\                                                                              
canoeing or kayaking & 14 & 0.0 & (1.5)& 0.0 & (5.3)& 0.0 & (2.6)& 0.0 & (2.8)& 0.0 & \textbf{(8.2)}& 0.0 & (3.9) \\                                  
catching or throwing baseball & 14 & \textbf{21.4} & \textbf{(27.0)}& 0.0 & (22.9)& 0.0 & (9.2)& 0.0 & (5.9)& 0.0 & (2.6)& 0.0 & (17.6) \\                                                                  
catching or throwing frisbee & 14 & \textbf{21.4} & (31.5)& \textbf{21.4} & \textbf{(42.7)}& 7.1 & (28.1)& 0.0 & (10.3)& 0.0 & (6.7)& \textbf{21.4} & (39.5) \\
clean and jerk & 13 & 15.4 & (25.3)& 38.5 & (47.7)& \textbf{46.2} & \textbf{(52.3)}& 23.1 & (43.0)& 30.8 & (47.5)& \textbf{46.2} & (50.1) \\                                                                
cleaning windows & 16 & \textbf{12.5} & \textbf{(17.0)}& 6.2 & (11.0)& 6.2 & (8.6)& 0.0 & (1.2)& 0.0 & (1.2)& 0.0 & (3.6) \\                                                                                
climbing a rope & 14 & 0.0 & (1.2)& 0.0 & (1.1)& 0.0 & \textbf{(9.5)}& 0.0 & (6.2)& 0.0 & (4.8)& 0.0 & (5.1) \\                                       
climbing ladder & 13 & 0.0 & (1.8)& 0.0 & (5.4)& \textbf{7.7} & \textbf{(11.4)}& 0.0 & (1.2)& 0.0 & (1.8)& 0.0 & (2.1) \\                                                                                   
deadlifting & 11 & 36.4 & (52.8)& 45.5 & (64.9)& 36.4 & (55.2)& 54.5 & (69.2)& 45.5 & (67.1)& \textbf{63.6} & \textbf{(75.5)} \\                                                                            
dribbling basketball & 18 & 5.6 & (11.6)& 22.2 & (31.5)& 50.0 & (60.6)& 44.4 & (49.4)& \textbf{61.1} & \textbf{(67.9)}& 27.8 & (46.9) \\                                                                    
drinking & 27 & 3.7 & (10.4)& 0.0 & (13.6)& 0.0 & \textbf{(13.9)}& 0.0 & (0.9)& 0.0 & (0.9)& \textbf{7.4} & (10.3) \\                                                                                       
driving car & 16 & 0.0 & (2.9)& 0.0 & (3.7)& \textbf{6.2} & (8.8)& 0.0 & (2.1)& 0.0 & (0.7)& \textbf{6.2} & \textbf{(9.4)} \\                                                                               
dunking basketball & 10 & 40.0 & (55.3)& \textbf{60.0} & \textbf{(64.3)}& 20.0 & (28.9)& 30.0 & (41.8)& 0.0 & (6.3)& 40.0 & (47.9) \\                                                                       
eating cake & 19 & 0.0 & (2.0)& 0.0 & \textbf{(2.9)}& 0.0 & (1.3)& 0.0 & (0.6)& 0.0 & (1.4)& 0.0 & (0.9) \\                                           
eating ice cream & 11 & 0.0 & (4.7)& 0.0 & (11.3)& 0.0 & (4.4)& 0.0 & (2.4)& 0.0 & (1.8)& \textbf{18.2} & \textbf{(21.5)} \\                                                                                
flying kite & 10 & \textbf{10.0} & \textbf{(14.4)}& 0.0 & (3.7)& 0.0 & (3.2)& 0.0 & (1.6)& 0.0 & (1.2)& 0.0 & (6.6) \\                                                                                      
golf driving & 16 & 12.5 & (19.7)& 31.2 & (44.0)& \textbf{62.5} & (69.5)& 50.0 & (57.8)& 37.5 & (47.5)& \textbf{62.5} & \textbf{(70.5)} \\                                                                  
hitting baseball & 15 & 6.7 & (18.5)& 13.3 & (23.4)& 0.0 & (17.7)& \textbf{20.0} & \textbf{(34.9)}& 6.7 & (16.0)& \textbf{20.0} & (34.1) \\                                                                 
hurdling & 10 & 0.0 & (13.5)& \textbf{20.0} & \textbf{(29.7)}& \textbf{20.0} & (29.2)& 0.0 & (9.8)& 0.0 & (11.0)& 10.0 & (23.3) \\                                                                          
juggling balls & 12 & 33.3 & (40.9)& 25.0 & (39.7)& 33.3 & (53.0)& \textbf{58.3} & \textbf{(60.3)}& 25.0 & (35.5)& 16.7 & (32.6) \\                                                                         
juggling soccer ball & 18 & 11.1 & (23.9)& 5.6 & (25.5)& \textbf{50.0} & \textbf{(61.6)}& 0.0 & (12.1)& 27.8 & (41.8)& 44.4 & (57.5) \\                                                                     
opening bottle & 9 & 0.0 & (1.7)& 0.0 & (4.7)& \textbf{11.1} & \textbf{(13.8)}& 0.0 & (1.0)& 0.0 & (0.8)& 0.0 & (6.9) \\                                                                                    
playing accordion & 11 & 0.0 & (4.7)& 9.1 & (18.5)& 9.1 & (11.2)& 0.0 & (6.7)& 0.0 & (5.0)& \textbf{27.3} & \textbf{(36.1)} \\                                                                              
playing basketball & 14 & 7.1 & (21.6)& \textbf{14.3} & \textbf{(35.5)}& 0.0 & (27.9)& 7.1 & (23.6)& 0.0 & (7.3)& 0.0 & (10.5) \\                                                                           
playing bass guitar & 13 & 0.0 & (5.2)& 7.7 & (12.4)& 7.7 & (20.3)& 0.0 & (6.0)& 0.0 & (3.2)& \textbf{15.4} & \textbf{(27.7)} \\                                                                            
playing guitar & 18 & \textbf{5.6} & (9.2)& \textbf{5.6} & (12.8)& \textbf{5.6} & (14.4)& 0.0 & (3.1)& 0.0 & (1.1)& \textbf{5.6} & \textbf{(14.9)} \\                                                       
playing piano & 17 & 0.0 & (9.6)& 11.8 & (18.7)& \textbf{17.6} & \textbf{(19.6)}& 0.0 & (6.8)& 5.9 & (11.2)& 11.8 & (13.5) \\                                                                               
playing saxophone & 13 & 0.0 & (2.7)& 0.0 & (6.3)& \textbf{7.7} & (9.1)& 0.0 & (3.7)& 0.0 & (4.0)& 0.0 & \textbf{(14.2)} \\                                                                                 
playing tennis & 19 & 5.3 & (7.9)& 10.5 & (15.1)& 21.1 & (35.0)& \textbf{31.6} & \textbf{(45.5)}& 5.3 & (20.4)& 21.1 & (34.5) \\                                                                            
playing trumpet & 14 & 0.0 & (8.0)& 21.4 & (25.1)& 7.1 & (14.0)& 0.0 & (14.2)& 0.0 & (12.7)& \textbf{35.7} & \textbf{(47.2)} \\                                                                             
playing violin & 20 & 10.0 & (15.3)& 10.0 & (26.0)& 5.0 & (15.2)& \textbf{25.0} & \textbf{(37.5)}& \textbf{25.0} & (34.7)& \textbf{25.0} & (36.5) \\                                                        
playing volleyball & 13 & 30.8 & (44.4)& 7.7 & (28.3)& \textbf{38.5} & \textbf{(52.9)}& 0.0 & (5.3)& 0.0 & (4.5)& 7.7 & (18.4) \\                                                                           
punching person (boxing) & 16 & 12.5 & (22.8)& 18.8 & (30.3)& \textbf{25.0} & \textbf{(31.3)}& 6.2 & (19.8)& 0.0 & (8.8)& 12.5 & (20.3) \\                                                                  
reading book & 10 & 0.0 & (1.8)& 0.0 & (6.0)& 0.0 & (3.5)& 0.0 & (2.1)& 0.0 & (2.3)& \textbf{10.0} & \textbf{(17.9)} \\                                                                                     
reading newspaper & 10 & 0.0 & (2.3)& 0.0 & (1.1)& 0.0 & (1.2)& 0.0 & (0.7)& 0.0 & (0.5)& 0.0 & \textbf{(3.1)} \\                                     
shooting basketball & 19 & \textbf{5.3} & (15.4)& \textbf{5.3} & \textbf{(20.2)}& \textbf{5.3} & (11.3)& \textbf{5.3} & (19.2)& 0.0 & (3.4)& \textbf{5.3} & (13.4) \\                                       
shooting goal (soccer) & 14 & 7.1 & (23.9)& 0.0 & (21.2)& 7.1 & (22.6)& 7.1 & (24.3)& 0.0 & (10.0)& \textbf{14.3} & \textbf{(29.8)} \\                                                                      
skiing (not slalom or crosscountry) & 10 & 0.0 & (4.1)& \textbf{20.0} & \textbf{(23.0)}& 0.0 & (1.5)& 0.0 & (1.1)& 0.0 & (1.4)& 0.0 & (2.0) \\                                                              
skiing slalom & 10 & 0.0 & (5.5)& 0.0 & (1.5)& 0.0 & (0.6)& 10.0 & (13.3)& \textbf{20.0} & \textbf{(20.8)}& 10.0 & (15.8) \\                                                                                
skipping rope & 12 & 41.7 & (53.6)& 41.7 & (58.5)& \textbf{75.0} & \textbf{(83.3)}& \textbf{75.0} & (81.3)& 0.0 & (8.8)& 50.0 & (61.6) \\                                                                   
smoking & 19 & 0.0 & (8.5)& \textbf{5.3} & \textbf{(14.8)}& 0.0 & (7.2)& 0.0 & (2.0)& 0.0 & (1.5)& \textbf{5.3} & (13.8) \\                                                                                 
surfing water & 10 & 0.0 & \textbf{(6.9)}& 0.0 & (2.9)& 0.0 & (6.6)& 0.0 & (4.2)& 0.0 & (3.6)& 0.0 & (2.6) \\                                         
sweeping floor & 11 & 0.0 & (1.9)& 0.0 & (1.7)& 0.0 & (1.4)& 0.0 & (1.0)& 0.0 & \textbf{(3.0)}& 0.0 & (0.9) \\                                        
sword fighting & 17 & 0.0 & (15.2)& \textbf{17.6} & \textbf{(36.1)}& 11.8 & (25.2)& 0.0 & (7.2)& 0.0 & (5.4)& 0.0 & (10.2) \\                                                                               
tying tie & 8 & 0.0 & (7.3)& 0.0 & (7.4)& \textbf{12.5} & \textbf{(21.5)}& 0.0 & (6.2)& 0.0 & (0.8)& 0.0 & (13.4) \\                                                                                        
walking the dog & 15 & \textbf{6.7} & \textbf{(11.7)}& 0.0 & (4.9)& 0.0 & (2.8)& 0.0 & (1.6)& 0.0 & (2.0)& 0.0 & (3.7) \\                                                                                   
writing & 13 & 0.0 & (1.8)& 0.0 & (3.0)& 0.0 & (2.9)& 0.0 & (2.6)& 0.0 & (0.9)& \textbf{15.4} & \textbf{(18.5)} \\                                                                                          
\hline
\textbf{avg} (50 classes) & 713  & 8.6 & (15.6)& 11.8 & (21.1)& 12.6 & (20.7)& 9.0 & (15.4)& 5.8 & (11.3)& \textbf{14.2} & \textbf{(22.7)} \\                                                               
\hline

 \end{tabular}
 }
 \label{tab:results}
\end{table*}

\section{ Mimetics evaluation with Superclasses }
\label{app:superclass}

\begin{figure}
 \includegraphics[width=\linewidth]{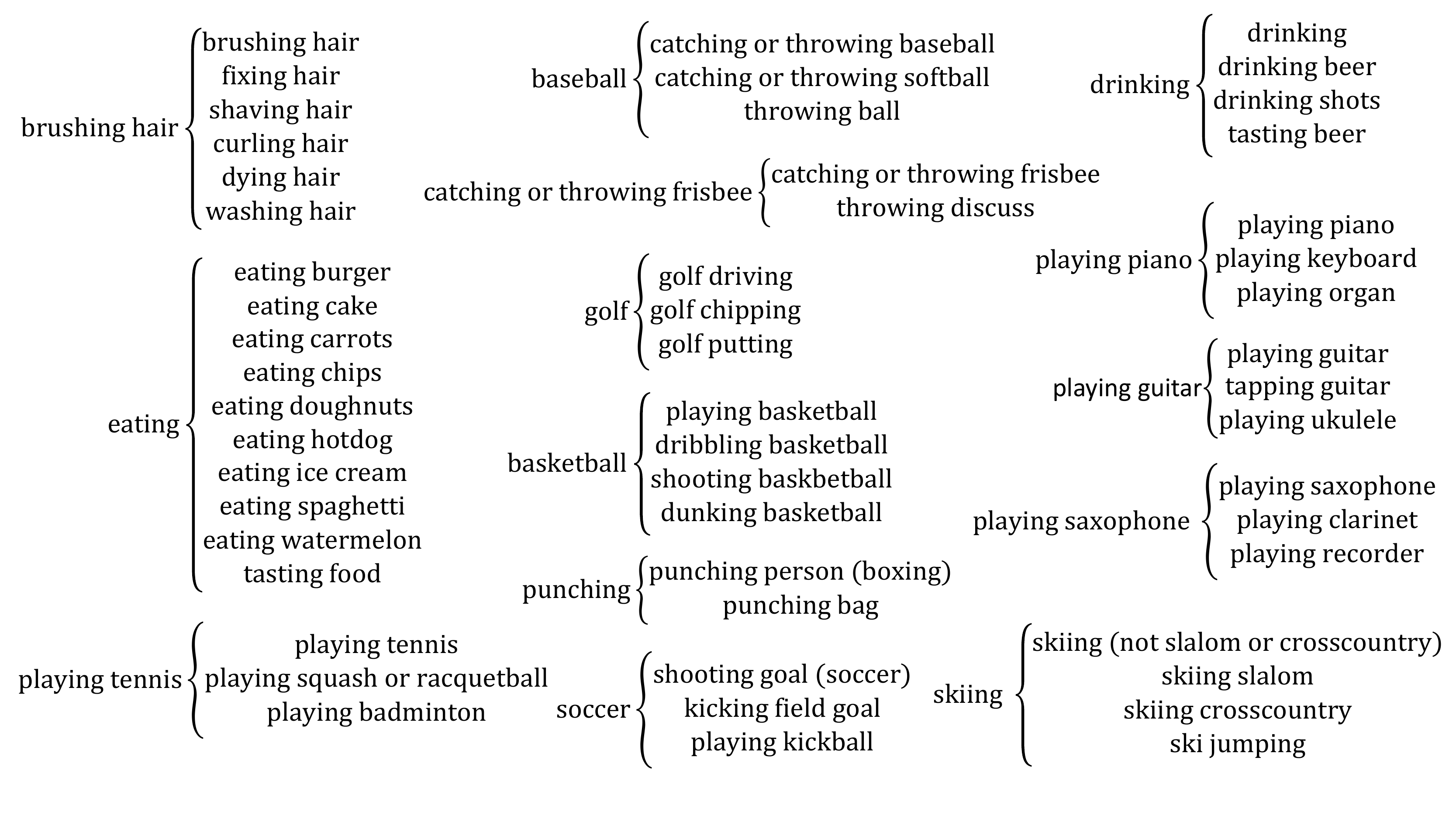}
 \caption{Superclasses employed in this paper. Classes not mentioned here are kept as is and considered as superclasses.}
 \label{fig:superclass}
\end{figure}

One hypothesis for the low performance of the different methods on Mimetics is the fact that actions are too fine-grained.
To evaluate this aspect, we aggregated several actions from the Kinetics dataset into some higher-level classes, denoted as superclasses.
Figure~\ref{fig:superclass} shows the actions that belong to a larger group. Classes not present in one of this group are kept as is, \ie, they form their own superclass.

For evaluation, we still consider models trained on the 400 classes from Kinetics, but at test time, when evaluating on Mimetics, we transformed the 400 class probabilities into superclass probabilities. 
As a side effect, the number of 50 classes from Mimetics is down to 45 superclasses.
Note that some elements in the superclasses do not belong to the subset of classes covered by Mimetics but are still used to sum the probabilities of a superclass from the Kinetics predictions.
Results are presented in Table~\ref{tab:superclass} (top part with 50 classes, bottom part with 45 classes).
We observe that the performance of all methods slightly increase compared to the evaluation with fine-grained classes (Table~\ref{tab:mimetics}) but the overall conclusion still holds.
Note in particular that  performances of pose-based methods show a larger increase, \eg +10\% top-5 accuracy for STGCN3D or +4\% for SIP-Net compared to +2\% for 3D-ResNeXt-101 RGB.

\begin{table}
\caption{Evaluation on Mimetics using superclasses.}
\begin{tabular}{l|rr|r}
   & top-1 & top-5 & mAP\\
\hline
\hline
RGB (3D-ResNeXt-101)                     &   10.9 &   22.5 &    18.3 \\
Flow (3D-ResNeXt-101)                    &   15.7 &   35.4 &    25.5 \\
RGB+Flow (late fusion)                   &   13.1 &   31.0 &    22.2 \\
STGCN (OpenPose)                         &   13.1 &   34.5 &    23.1 \\
\hline
STGCN2D                                  &    8.3 &   25.9 &    17.7 \\
STGCN3D                                  &    7.9 &   24.8 &    16.6 \\
SIP-Net                                  &   {\bf15.9} &  {\bf35.9} &  {\bf25.9} \\
 \end{tabular}
\label{tab:superclass}
\end{table}

\end{document}